%% file: main.tex
\documentclass[]{fairmeta}
\usepackage{pifont}

\usepackage{amsmath}
\usepackage{amssymb}
\usepackage{makecell} 
\usepackage{atbegshi}
\usepackage{mathtools}
\usepackage{amsthm}
\usepackage{enumitem}
\usepackage{algorithm}
\usepackage{wrapfig}
\usepackage{wrapfig}
\usepackage{algpseudocode}
\usepackage{booktabs}
\usepackage{array}
\usepackage{booktabs}
\usepackage{array}
\usepackage{makecell}

\setcitestyle{authoryear,round,aysep={,},yysep={;}}

\usepackage{tabularx}
\usepackage{makecell}
\usepackage{booktabs}
\usepackage[most]{tcolorbox}
\usepackage{xcolor}
\usepackage{fvextra}
\usepackage{xurl}
\usepackage{hyperref}

\renewcommand{\cite}[1]{(\citealp{#1})}
\renewcommand{\citep}[1]{(\citealp{#1})}
\usepackage{eso-pic}

\newcolumntype{L}[1]{>{\raggedright\arraybackslash}m{#1}}
\newcolumntype{C}[1]{>{\centering\arraybackslash}m{#1}}

\usepackage[most]{tcolorbox} 
\newtcolorbox{rolebox}{enhanced, breakable, colback=black!4!white, colframe=black!70,
  boxrule=0.6pt, arc=2.5mm, left=2.5mm, right=2.5mm, top=1.5mm, bottom=1.5mm,
  before skip=6pt, after skip=8pt}

\newtcolorbox{templatebox}[1]{
    breakable,
    enhanced,
    colback=white,
    colframe=gray!80!black,
    colbacktitle=gray!80!black,
    coltitle=white,
    fonttitle=\bfseries,
    title=#1,
    arc=3mm,
    boxrule=1pt,
    drop fuzzy shadow={gray!50!white},
    left=5mm,
    right=5mm,
    top=3mm,
    bottom=3mm
}

\renewcommand{\affiliation}[2][]{%
  \addtolist[#1]{#2}{\affiliationlist}{\affiliationformat}{\quad}%
}

\title{
LongHorizon-Harness: Advancing Long-Horizon Agents for Real-World Tasks
}

\author[*]{Ziyu Ma}
\author[*]{Hailang Huang}
\author[*]{Shun Zou}
\author[*\dagger]{Yong Wang}
\author[]{Shidong Yang}
\author[]{Yiming Hu}
\author[]{Fei Wei}
\author[]{Xiangxiang Chu}

\affiliation[ ]{AMAP, Alibaba Group}
\affiliation[*]{Equal contribution}
\affiliation[\dagger]{Project lead}

\abstract{%
Large language model (LLM) agents increasingly undertake long-horizon tasks that require sustained reasoning, tool use, and revision across many interdependent steps. However, existing agent harnesses maintain task execution, task state, and completion assessment within a growing context, making the state difficult to track and allowing incorrect self-assessments to propagate into later decisions. We reformulate long-horizon execution as a task-state management problem and propose \textbf{LongHorizon-Harness}, which maintains the task state explicitly outside execution and updates it only with facts independently verified from the environment. Its \emph{Manage-Execute-Audit} (MEA) loop uses a manager to maintain the task state and determine the next subtask, a fresh-context executor to perform it, and a read-only auditor to verify the resulting environment state before the next round. A lightweight \emph{AgentAdapter} supports interchangeable model and harness backends without modifying their native agent loops. LongHorizon-Harness improves Qwen~3.7-Plus from 51.8\% to 80.7\% on WeaveBench, from 69.7\% to 77.2\% on Terminal-Bench~2.1, and from 2.8\% to 8.3\% on OSWorld~2.0. It also raises Claude Opus~4.7 from 20.0\% to 34.3\% on an OSWorld~2.0 subset, demonstrating consistent gains across models, harnesses, and interaction domains.

}

\metadata[Github]{\url{https://github.com/AMAP-ML/LongHorizon-Harness}}

\metadata[Website]{\url{https://lh-harness.pages.dev}}

\begin{document}
\maketitle

\begin{figure*}[ht]
    \centering
    \newlength{\teaserheight}
    \setlength{\teaserheight}{0.36\linewidth}
    \includegraphics[height=\teaserheight]{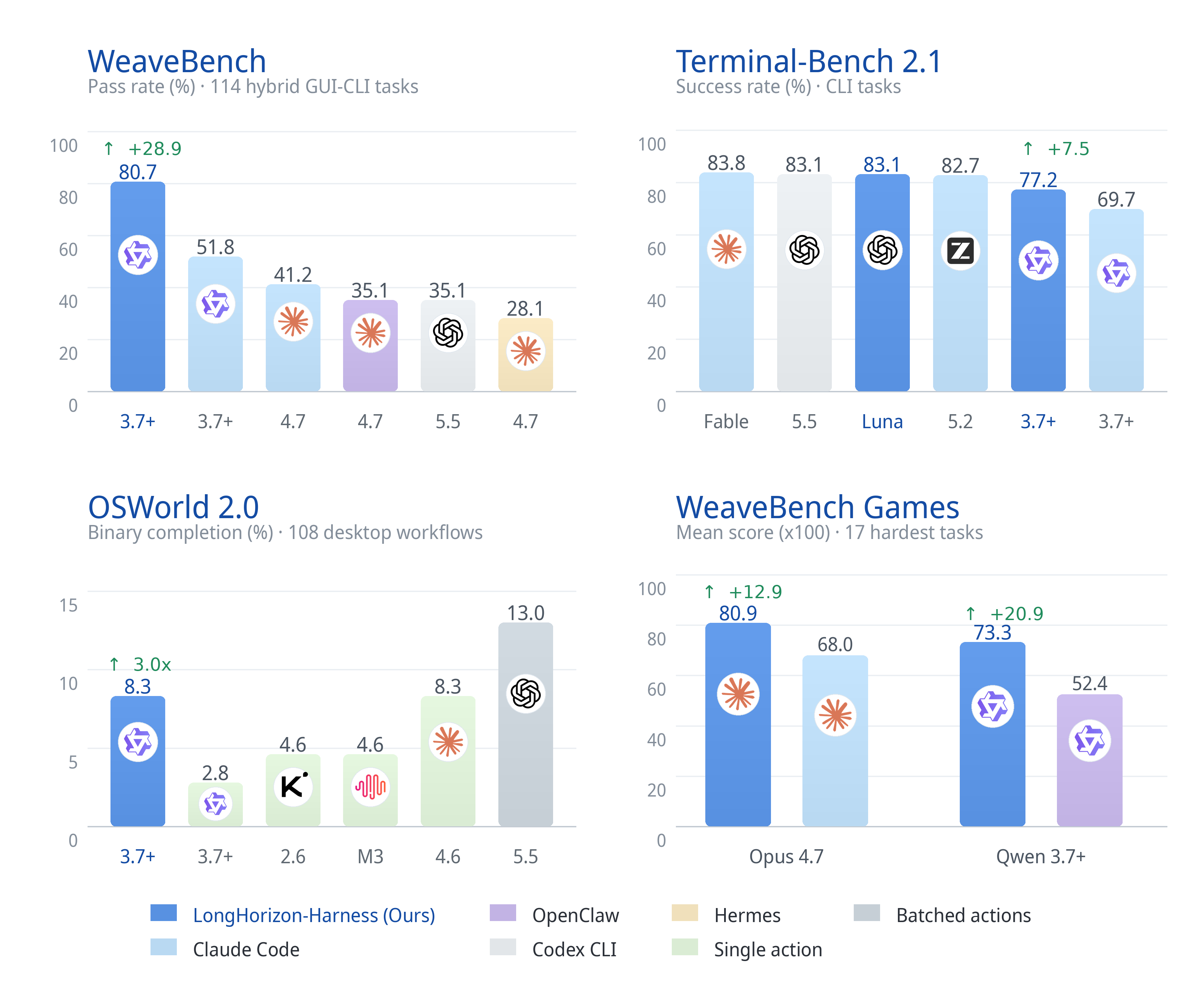}\hfill
    \includegraphics[height=\teaserheight]{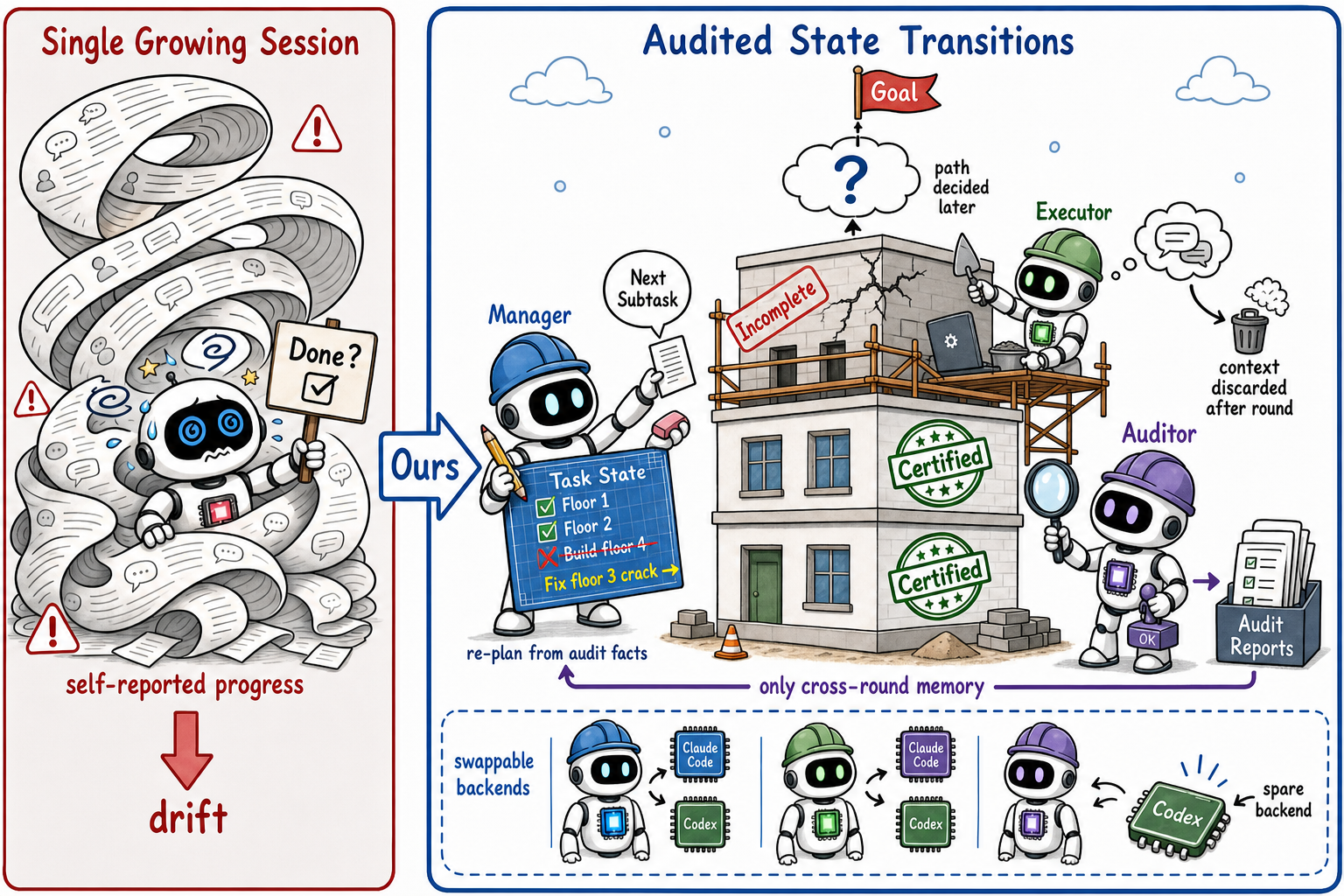}
    \vspace{-0.25cm}
    \caption{\textbf{Left: LongHorizon-Harness improves long-horizon execution across benchmarks and backbones.}
    With the same backbone and execution backend, it lifts WeaveBench PassRate from 51.8\% to 80.7\%, Terminal-Bench~2.1 from 69.7\% to 77.2\%, and OSWorld~2.0 binary completion by $3.0\times$, and the gains transfer from Qwen~3.7-Plus to Claude Opus~4.7.
    \textbf{Right: Audited state transitions.}
    Instead of one continuously growing session that judges its own progress, a \emph{manager} re-plans the next subtask from audited facts, a fresh-context \emph{executor} performs it, and a read-only \emph{auditor} certifies what actually changed in the environment. Audit reports are the only cross-round memory, and interchangeable backends (e.g., Claude Code, Codex) serve each role.}
    \vspace{-0.2cm}
    \label{fig:teaser}
\end{figure*}

\section{Introduction}
\label{sec:intro}


Over the past few years, large language models (LLMs) have evolved from conversational models into the decision-making core of autonomous agents for software engineering~\cite{sweagent,openhands,ren2026saasbench,ding2025nl2repo}, general-purpose assistance~\cite{claudecode,codex,openai_chatgpt_agent,anthropic_cowork}, scientific discovery~\cite{scientificdiscovery,wei2025ai}, computer use~\cite{anthropic_claude_use,openai_operator,google_gemini_cu,sager2026comprehensive,zhou2026colorbrowseragent,zheng2026code2world}, and multimodal interaction~\cite{multimodal_agent,agents}. Across these applications, agents increasingly face \emph{long-horizon execution}, which requires repeated reasoning, tool use, observation, and revision over many interdependent steps, sometimes across multiple context windows or sessions. The length of tasks that agents can complete increasingly determines how much work can be delegated to them. METR~\cite{longtasks} reports that the task-completion horizon of advanced agents has doubled roughly every seven months, with the trend accelerating to about four months for recent models. Frontier coding agents can already sustain hours-long work on a single project~\cite{codex,claudecode,yang2026programbench}. However, a longer horizon does not by itself make execution reliable~\cite{dong2026longhorizon}.

The difficulty of long-horizon execution lies not in any individual step, but in sustaining coherent progress across a long sequence of interdependent actions. Across systems and domains, three challenges consistently emerge: \textit{(i)~Compounding errors and goal drift.} Errors in earlier actions or decisions accumulate along the trajectory, distort subsequent choices, and gradually steer the agent away from its original objective~\cite{ale}. \textit{(ii)~Context rot.} As the interaction history grows, relevant information becomes increasingly difficult to retrieve and use, and agent performance can degrade sharply once context utilization crosses a critical threshold~\cite{contextlimit1,contextlimit2}. \textit{(iii)~Task-state loss.} Long-horizon tasks are difficult to complete without an accurate and up-to-date task state (i.e., the requirements to satisfy, actions already completed, artifacts produced, and facts discovered from the environment), but agents often fail to recover, retain, and update this state throughout task execution.

Extensive efforts have strengthened agents from both the model and harness sides. Frontier models continue to scale in size, extend their context windows, and acquire stronger coding and agentic capabilities by training on large-scale, high-quality data~\cite{swerl,chu2025gpg,li2025adacurl,ma2026skillclaw, ji2025tree, yang2026coevolve}. In parallel, agent harnesses such as Claude Code~\cite{claudecode}, Codex CLI~\cite{codex}, and OpenClaw~\cite{openclaw} have become the standard system layer for organizing prompting, tool use, context management, and multi-step execution around the model. For long-horizon tasks, these harnesses already support planning, task decomposition, tool use, and subagents that execute or review work in isolated contexts. However, existing harnesses still face two structural limitations: \textit{(i)~Task execution and task-state management share the same growing context.} The agent uses the same context to execute the task and maintain its task state, while the growing execution history makes task state increasingly difficult to track. \textit{(ii)~Task execution and completion assessment remain coupled.} The agent performs each subtask and judges whether it has been completed, with an incorrect judgment potentially being recorded as part of the task state and used as a premise for subsequent decisions.

To address these limitations, we propose \textbf{LongHorizon-Harness}, a framework that organizes long-horizon execution as a sequence of independently audited task-state transitions. Our key idea is to maintain the task state as an explicit record outside task execution, update it only with facts independently verified from the environment, and derive each next subtask from the current record and the original goal. Specifically, LongHorizon-Harness follows a \emph{Manage-Execute-Audit} (MEA) loop as shown in Fig.~\ref{fig:teaser}. A manager reads the current task state and defines one subtask with its dependencies, constraints, and acceptance criteria. An executor performs only this subtask in a fresh context, while a read-only auditor independently inspects the environment to determine what changed, what was completed, and what remains unmet~\cite{agentasjudge,llmasjudge}. The manager updates the task state from the audit result and begins the next round, while the executor's interaction history is discarded after each round so that only compact, verified task state persists across the task. Through a lightweight \emph{AgentAdapter}, LongHorizon-Harness preserves the native agent loops~\cite{react} of existing systems and supports interchangeable backends for all three roles, spanning models such as Claude Opus, GPT, and Qwen and harnesses such as Codex CLI~\cite{codex}, Claude Code~\cite{claudecode}, OpenClaw~\cite{openclaw}, and Hermes Agent~\cite{hermes}.

We evaluate LongHorizon-Harness on three recent long-horizon benchmarks, WeaveBench~\cite{weavebench}, OSWorld~2.0~\cite{osworld2}, and Terminal-Bench~2.1~\cite{terminalbench}, which together cover cross-interface coordination, desktop workflows, and challenging command-line tasks. Under matched model backends and evaluation protocols, LongHorizon-Harness raises WeaveBench PassRate from 51.8\% to \textbf{80.7\%} using Qwen~3.7-Plus and Claude Code, nearly doubling the strongest officially reported result of 41.2\% obtained by Claude Opus~4.7 with Claude Code. On Terminal-Bench~2.1, it improves performance from 69.7\% to \textbf{77.2\%} with Qwen~3.7-Plus. On the full OSWorld~2.0 benchmark, it raises binary completion from 2.8\% to \textbf{8.3\%} with Qwen~3.7-Plus; on a 34-task subset, it improves the result from 20.0\% to \textbf{34.3\%} with Claude Opus~4.7. Our contributions are summarized as follows:

\begin{itemize}[nosep]
    \item We reformulate long-horizon execution from a single growing trajectory into a task-state management problem. The task state is maintained explicitly outside execution, updated only with independently verified facts, and used to determine each next subtask under the original goal.

    \item We design LongHorizon-Harness, which realizes this principle through a \emph{Manage-Execute-Audit} loop. In each round, the manager defines one subtask from the current task state, the executor performs it in a fresh context, and the auditor independently inspects the environment before the manager updates the state and begins the next round.

    \item We evaluate LongHorizon-Harness on WeaveBench, OSWorld~2.0, and Terminal-Bench~2.1. With Qwen~3.7-Plus, it improves WeaveBench PassRate from 51.8\% to 80.7\%, Terminal-Bench~2.1 from 69.7\% to 77.2\%, and OSWorld~2.0 binary completion from 2.8\% to 8.3\%, while also delivering consistent gains with Claude Opus~4.7.
\end{itemize}

\begin{figure*}[t]
    \centering
    \includegraphics[width=\linewidth]{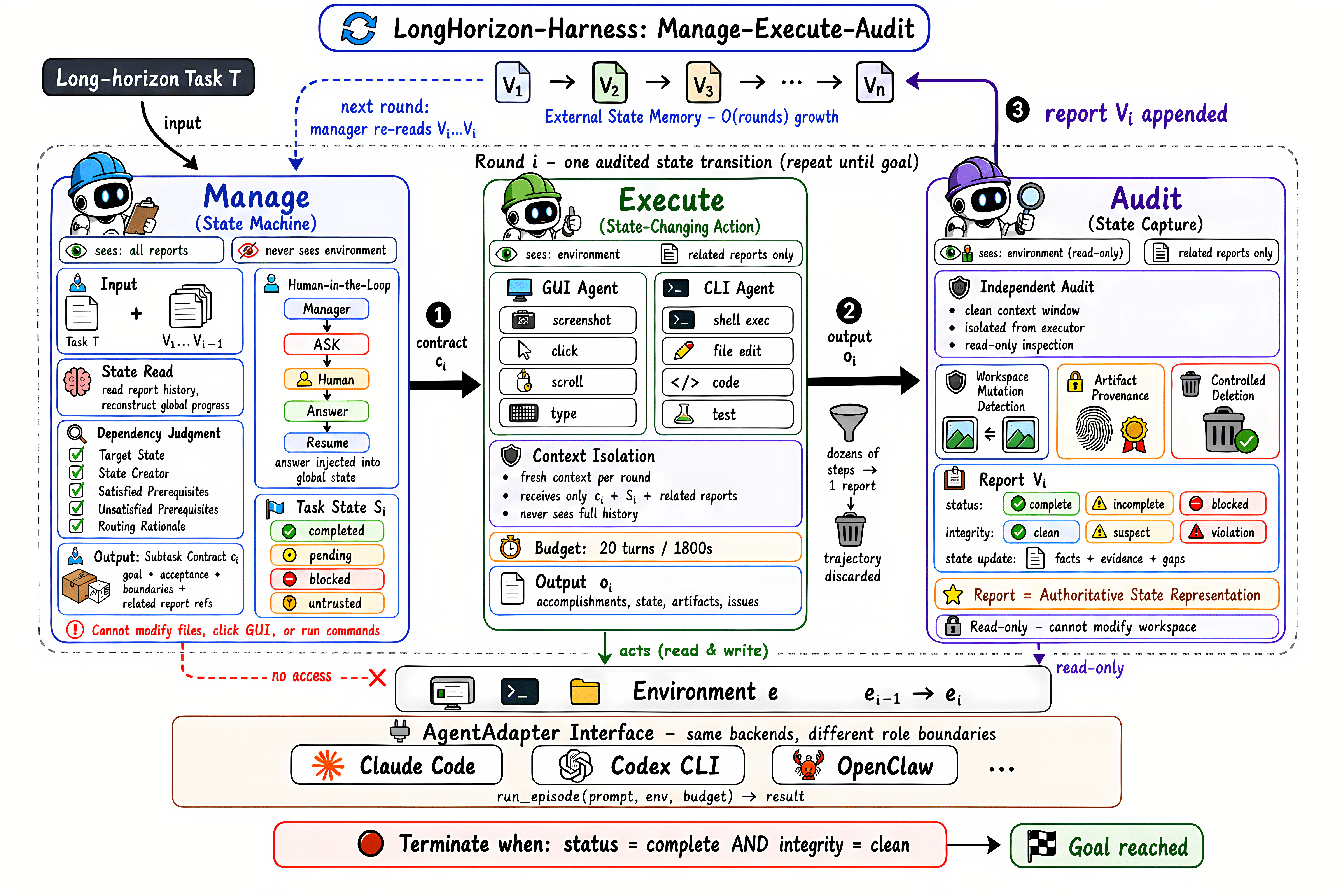}
    \vspace{-0.6cm}
    \caption{\textbf{Overview of LongHorizon-Harness.}
    LongHorizon-Harness processes a long-horizon task $\mathcal{T}$
    through repeated \emph{Manage-Execute-Audit} rounds, shown by the
    dashed box and numbered flow \ding{182}--\ding{184}. The
    \emph{manager} reads task state $S_i$ and constructs subtask
    contract $c_i$, which specifies the goal, acceptance criteria,
    boundary constraints, and relevant prior evidence (\ding{182}).
    The manager may instead request user information or authorization
    through the \texttt{ask} route. The selected GUI or CLI
    \emph{executor} performs the subtask in a fresh, budget-bounded
    context and modifies the environment (\ding{183}). The
    \emph{auditor} independently inspects the resulting environment
    through read-only tools and produces audit report $v_i$
    (\ding{184}). The manager uses $v_i$ to update the task state
    before the next round.}
    \vspace{-0.2cm}
    \label{fig:pipeline}
\end{figure*}

\input{sec/method}
\input{sec/experiment}
\section{Conclusion}
\label{sec:conclusion}

In this work, we introduce LongHorizon-Harness, a general framework for long-horizon agent execution that separates task-state management from environment interaction through a Manage--Execute--Audit loop. LongHorizon-Harness maintains progress as explicit, audited task state, executes each subtask in a fresh context, and carries only independently verified outcomes across rounds. Experiments on WeaveBench, OSWorld~2.0, and Terminal-Bench~2.1 show consistent improvements across hybrid GUI--CLI workflows, professional desktop tasks, pure command-line environments, and different model backbones. These results demonstrate that long-horizon agent capability is determined not only by the underlying model, but also by the harness that organizes, verifies, and converts its local capabilities into end-to-end task completion.

\clearpage
\bibliographystyle{plainnat}
\bibliography{ref}

\input{sec/appendix}

\end{document}

%% file: sec/method.tex
\section{Method}

\subsection{Overview}
\label{sec:method:overview}

Given a long-horizon task $\mathcal{T}$ and a computer environment,
LongHorizon-Harness executes the task through a sequence of dynamically
determined rounds rather than a single continuously growing session.
The harness maintains an explicit task state outside task execution and
advances it only with evidence independently verified from the
environment. Across rounds, only the task state and its supporting audit
reports persist; the executor's raw interaction trajectory is
discarded after each round. Fig.~\ref{fig:pipeline} provides an overview
of the framework.

Each round follows a \emph{Manage-Execute-Audit} (MEA) loop. Let $S_i$
denote the task state available at the beginning of round $i$,
$e_{i-1}\in\mathcal{E}$ the current environment state, and
$V_{i-1}=(v_1,\ldots,v_{i-1})$ the accumulated audit reports. The
manager constructs a bounded subtask contract $c_i$. A fresh-context
executor performs the contract, transforms the environment from
$e_{i-1}$ to $e_i$, and returns an execution report $o_i$. A separate
auditor then inspects $e_i$ through read-only tools and produces audit
report $v_i$. The manager incorporates $v_i$ into the next task state
$S_{i+1}$ before determining whether another round is required. The loop
ends when the audited state satisfies the original task, no
permitted subtask can advance the remaining requirements, user input is
required, or the round budget is exhausted.

\subsection{Manager}
\label{sec:method:manager}

The manager owns the persistent task state and determines how the task
should proceed. It has access to the original task $\mathcal{T}$, the
current task state, and all accumulated audit reports, but has no direct
interface to the computer environment. It cannot observe application
state, inspect workspace contents, invoke GUI or CLI tools, or modify
the environment. Its decisions are based entirely on the task state and
environment evidence recorded by the auditors.

After round $i$, the manager updates the task state and produces the
next control decision:
\begin{equation}
\label{eq:manager}
(S_{i+1},q_{i+1},c_{i+1})
=
\Phi_{\mathrm{mgr}}
\left(
\mathcal{T},
S_i,
V_i
\right),
\end{equation}
where $V_i=(v_1,\ldots,v_i)$ and $q_{i+1}$ is one of
\textsc{execute}, \textsc{done}, \textsc{blocked}, and \textsc{ask}.
The contract $c_{i+1}$ is returned only when further execution is
required.

\paragraph{Task-state update.}
The task state is a structured collection of task-relevant records:
a \textit{requirement} represents an objective or constraint derived
from the original task, an \textit{artifact} represents an output
created or modified during execution, and a \textit{fact} records
environment information needed by subsequent rounds. Each record is
marked as \textit{completed}, \textit{pending}, \textit{blocked}, or
\textit{untrusted}, and retains references to the audit evidence
supporting its current status. The initial state $S_1$ is constructed
from $\mathcal{T}$ with its requirements marked as pending. After each
round, the manager applies the verified findings in $v_i$ to $S_i$,
adding or updating the corresponding records while leaving unresolved
ones pending, blocked, or untrusted as appropriate. Executor claims do
not directly change the persistent state: a record is marked as
completed only when supported by clean audit evidence. 

\paragraph{Next-subtask construction.}
The manager compares $S_{i+1}$ against the original task, selects an
unresolved objective that can be advanced from the current state, and
checks its dependencies and prerequisites. It then constructs a bounded
contract $c_{i+1}$ specifying the immediate goal, acceptance criteria,
boundary constraints, and the task-state records and prior audit reports
relevant to execution and verification. The contract is routed to a GUI
or CLI executor according to the primary environment transition it
requires. The manager returns \textsc{done} when the audited state
satisfies $\mathcal{T}$ without unresolved integrity violations,
\textsc{blocked} when no permitted action can advance the remaining
requirements, and \textsc{ask} when progress requires user information
or authorization; otherwise, it returns \textsc{execute} together with
$c_{i+1}$.

\subsection{Executor}
\label{sec:method:executor}

The executor performs the contract selected by the manager and is the
only role permitted to intentionally modify the environment. In round
$i$, it receives the original task $\mathcal{T}$, the current task
state $S_i$, the subtask contract $c_i$, and only the prior audit
reports referenced by the contract. It transforms the environment from
$e_{i-1}$ to $e_i$:
\begin{equation}
\label{eq:executor}
(e_i,o_i)
=
\Phi_{\mathrm{exec}}
\left(
\mathcal{T},
S_i,
c_i;
e_{i-1}
\right),
\end{equation}
where $o_i$ summarizes the actions performed, resulting state,
artifacts produced or modified, and issues encountered during
execution. The report $o_i$ describes the executor's outcome but does
not establish that the contract has been completed.

\paragraph{Fresh-context execution.}
Each executor invocation runs as a fresh, budget-bounded episode
containing only the information supplied for the current round. It does
not receive the raw interaction trajectories of earlier rounds. Within
the episode, the executor may perform multiple cycles of planning,
environment interaction, observation, and revision. When the episode
ends, its raw trajectory and internal reasoning are discarded; only
$o_i$ is forwarded for auditing.

\paragraph{GUI and CLI capability boundaries.}
GUI and CLI executors operate through different environment interfaces.
The GUI executor receives screen-oriented capabilities such as taking
screenshots, clicking, scrolling, and entering text, and is responsible
for transitions centered on application and interface state. The CLI
executor receives capabilities such as shell execution, file editing,
coding, and testing, and is responsible for transitions centered on
workspace, process, and program state.

The harness exposes only the environment interface and tool set
assigned to the selected executor role. Capabilities outside that role
are unavailable unless explicitly provided by its configuration. This
separates responsibility for different classes of state-changing
actions while allowing the manager to select the interface appropriate
for each contract.

\paragraph{Backend execution.}
Executors are instantiated through a common agent-adapter interface.
Given a contract, a role-specific environment interface, and an
execution budget, the adapter launches an existing backend such as
Claude Code, Codex CLI, or OpenClaw as one bounded episode. The backend
retains its native planning and tool-use loop and may normally invoke
the shell, edit files, write code, run tests, or interact with
applications when those capabilities are exposed by its assigned role.
The harness does not replace the backend's internal execution process;
it controls the supplied context, available tools, environment
permissions, execution budget, and returned report.

\subsection{Auditor}
\label{sec:method:auditor}

The auditor independently verifies the environment state produced by
the executor. After execution, it receives the original task
$\mathcal{T}$, the task state $S_i$, the contract $c_i$, the prior
audit reports referenced by the contract, and the executor report
$o_i$. It does not receive the executor's raw interaction trajectory or
internal reasoning. The auditor inspects the resulting environment
$e_i$ and produces
\begin{equation}
\label{eq:auditor}
v_i
=
\Phi_{\mathrm{aud}}
\left(
\mathcal{T},
S_i,
c_i,
o_i;
e_i
\right),
\end{equation}
where $v_i$ is appended to the persistent audit history and returned to
the manager before the next round.

\paragraph{Independent environment inspection.}
The auditor starts from a fresh context that excludes the executor's
raw interaction trajectory and internal reasoning. It may use $o_i$ to
locate relevant files, windows, logs, processes, or other outputs, but
determines completion by independently comparing the resulting
environment against the goal, acceptance criteria, and boundary
constraints in $c_i$. A GUI auditor examines application and screen
state through observation-oriented interactions, whereas a CLI auditor
uses non-mutating commands and inspection tools to examine files,
metadata, logs, processes, tests, and workspace state. In both cases,
audit conclusions must be supported by evidence obtained directly from
the environment rather than by the executor's completion claim.

\paragraph{Read-only authority.}
The auditor may change its observation view when necessary for
inspection but cannot modify task-relevant environment state. It cannot
create, edit, overwrite, move, or delete protected artifacts, execute
state-changing commands, or perform GUI actions that alter the result
under inspection. The harness monitors task-relevant workspace and
artifact state throughout auditing; any detected mutation is recorded
as an integrity violation, and the resulting report cannot support a
completed task-state record.

\paragraph{Audit-result construction.}
The audit report $v_i$ records three classes of findings. First, it
assigns a completion status of \textit{complete},
\textit{incomplete}, or \textit{blocked} by evaluating the contract's
acceptance criteria. Second, it assigns an integrity status of
\textit{clean}, \textit{suspect}, or \textit{violation} by checking
workspace mutations, artifact validity and provenance, and relevant
deletion constraints. Third, it records the task-state updates supported
by the inspection, including verified facts, supporting evidence, and
remaining gaps.

The auditor may propose changes to requirement, artifact, and fact
records, but the manager determines how these findings are incorporated
into $S_{i+1}$. Consequently, $o_i$ remains an unverified execution
summary, whereas $v_i$ provides the environment-grounded evidence that
may advance the persistent task state.

%% file: sec/experiment.tex
\section{Experiments}
\label{sec:experiment}

\newcommand{\g}[1]{\textcolor{gray}{#1}}

\begin{table*}[t]
\centering
\footnotesize
\setlength{\tabcolsep}{3pt}
\renewcommand{\arraystretch}{1.15}
\caption{\textbf{Results on WeaveBench.}
Gray rows denote official results reported by \cite{weavebench}, using the best reported thinking mode for each backbone; models other than GPT-5.5 and Claude Opus~4.7 were evaluated only with \textsc{OpenClaw}.
Black rows denote our runs with Qwen~3.7-Plus.
Our runs use root privileges inside the task virtual machine, whereas the official results use a regular user account and are therefore included as reference points rather than matched comparisons.
PR denotes full-task PassRate~(\%), and Overall denotes the mean score over all 114 tasks.
The remaining columns report PassRate across the eight benchmark domains.}

\label{tab:weavebench}
\begin{tabular}{l l cc | cccccccc}
\toprule
\textbf{Model} & \textbf{Harness} & \textbf{PR}$\uparrow$ & \textbf{Overall}$\uparrow$ & \textbf{DSK} & \textbf{DOC} & \textbf{GAM} & \textbf{WEB} & \textbf{DAV} & \textbf{OPS} & \textbf{SPA} & \textbf{DES} \\
\midrule
\multirow{4}{*}{\g{Claude Opus 4.7}}
& \g{\textsc{Claude Code}}  & \g{41.2} & \g{0.532} & \g{55.6} & \g{47.1} & \g{23.5} & \g{53.3} & \g{23.1} & \g{50.0} & \g{33.3} & \g{40.0} \\
& \g{\textsc{OpenClaw}}     & \g{35.1} & \g{0.482} & \g{55.6} & \g{29.4} & \g{23.5} & \g{66.7} & \g{15.4} & \g{41.7} & \g{16.7} & \g{20.0} \\
& \g{\textsc{Hermes Agent}} & \g{28.1} & \g{0.516} & \g{33.3} & \g{47.1} & \g{11.8} & \g{26.7} & \g{30.8} & \g{50.0} & \g{8.3}  & \g{10.0} \\
& \g{\textsc{Codex CLI}}    & \g{13.2} & \g{0.378} & \g{16.7} & \g{11.8} & \g{11.8} & \g{6.7}  & \g{7.7}  & \g{25.0} & \g{16.7} & \g{10.0} \\
\midrule
\multirow{4}{*}{\g{GPT-5.5}}
& \g{\textsc{Codex CLI}}    & \g{35.1} & \g{0.499} & \g{38.9} & \g{29.4} & \g{23.5} & \g{53.3} & \g{15.4} & \g{50.0} & \g{58.3} & \g{10.0} \\
& \g{\textsc{OpenClaw}}     & \g{33.3} & \g{0.466} & \g{38.9} & \g{35.3} & \g{35.3} & \g{21.4} & \g{23.1} & \g{38.5} & \g{33.3} & \g{40.0} \\
& \g{\textsc{Hermes Agent}} & \g{31.6} & \g{0.466} & \g{55.6} & \g{29.4} & \g{35.3} & \g{40.0} & \g{7.7}  & \g{25.0} & \g{25.0} & \g{20.0} \\
& \g{\textsc{Claude Code}}  & \g{14.9} & \g{0.299} & \g{33.3} & \g{11.8} & \g{11.8} & \g{0.0}  & \g{15.4} & \g{16.7} & \g{25.0} & \g{0.0}  \\
\midrule
\g{GPT-5.4}           & \g{\textsc{OpenClaw}} & \g{22.8} & \g{0.465} & \g{55.6} & \g{35.3} & \g{5.9}  & \g{0.0} & \g{23.1} & \g{23.1} & \g{8.3} & \g{20.0} \\
\g{GPT-5.3-codex}     & \g{\textsc{OpenClaw}} & \g{18.4} & \g{0.456} & \g{33.3} & \g{23.5} & \g{29.4} & \g{0.0} & \g{7.7}  & \g{16.7} & \g{8.3} & \g{20.0} \\
\g{GPT-5.2-codex}     & \g{\textsc{OpenClaw}} & \g{6.1}  & \g{0.321} & \g{5.6}  & \g{11.8} & \g{0.0}  & \g{0.0} & \g{15.4} & \g{16.7} & \g{0.0} & \g{0.0}  \\
\g{GPT-5.1-codex}     & \g{\textsc{OpenClaw}} & \g{1.8}  & \g{0.226} & \g{0.0}  & \g{5.9}  & \g{0.0}  & \g{0.0} & \g{7.7}  & \g{0.0}  & \g{0.0} & \g{0.0}  \\
\g{Gemini 3.1 pro}    & \g{\textsc{OpenClaw}} & \g{1.8}  & \g{0.223} & \g{0.0}  & \g{0.0}  & \g{0.0}  & \g{0.0} & \g{0.0}  & \g{8.3}  & \g{8.3} & \g{0.0}  \\
\g{Qwen3.5-397B-A17B} & \g{\textsc{OpenClaw}} & \g{0.9}  & \g{0.318} & \g{0.0}  & \g{0.0}  & \g{0.0}  & \g{0.0} & \g{0.0}  & \g{8.3}  & \g{0.0} & \g{0.0}  \\
\g{Qwen3-VL-8B-Think} & \g{\textsc{OpenClaw}} & \g{0.9}  & \g{0.092} & \g{0.0}  & \g{0.0}  & \g{0.0}  & \g{0.0} & \g{8.3}  & \g{0.0}  & \g{0.0} & \g{0.0}  \\
\g{GUI-Owl-1.5-32B}   & \g{\textsc{OpenClaw}} & \g{0.0}  & \g{0.065} & \g{0.0}  & \g{0.0}  & \g{0.0}  & \g{0.0} & \g{0.0}  & \g{0.0}  & \g{0.0} & \g{0.0}  \\
\midrule
\multirow{2}{*}{Qwen 3.7-Plus}
& Claude Code & 51.8 & 0.702 & 83.3 & 76.5 & 29.4 & 46.7 & 53.8 & 66.7 & 16.7 & 20.0 \\
& \textbf{LongHorizon-Harness (Claude Code)} & \textbf{80.7} & \textbf{0.835} & \textbf{88.9} & \textbf{100.0} & \textbf{58.8} & \textbf{73.3} & \textbf{84.6} & \textbf{91.7} & \textbf{66.7} & \textbf{80.0} \\
\bottomrule
\end{tabular}
\end{table*}

\subsection{Experimental Setup}
\label{sec:exp-setup}
\paragraph{Compared configurations.}
We use Qwen-3.7-Plus as the primary backbone model, and additionally evaluate Claude Opus-4.7 to assess the generality across backbone models. For each backbone model, we compare the original Claude Code evaluation framework~\cite{claudecode} with \textbf{LongHorizon-Harness}, where the latter uses Claude Code as its execution backend. Unless otherwise specified, the manager, executor, and auditor are instantiated with the same backbone model, ensuring that the comparison reflects the effect of task-state management rather than differences in model capability.

\paragraph{Implementation details.}
LongHorizon-Harness integrates existing agent backends through a unified AgentAdapter interface. Each role is assigned an independent execution budget: the executor is limited to 1800 seconds per round, while both the manager and the auditor are limited to 300 seconds. We set the maximum number of MEA rounds to $N_{\max}=25$. 

\paragraph{Benchmarks and metrics.}
We evaluate LongHorizon-Harness on three recent benchmarks that cover complementary forms of long-horizon execution.

\textbf{WeaveBench}~\cite{weavebench} consists of 114 tasks that require coordinated GUI and CLI interactions within the same workflow. We follow the standard WeaveBench evaluation protocol, using the task definitions, workspaces, runtime assets, and judge templates released by the official benchmark for all tasks. We report PassRate, defined as the percentage of fully passed tasks, and Overall, defined as the average score across all tasks. In addition, we report PassRate for each of the eight domains covered by the benchmark.

\textbf{OSWorld 2.0}~\cite{osworld2} consists of 108 desktop workflow tasks, with a median human completion time of approximately 1.6 hours. We use the official osworld-v2-2026.06.24 release and its standard Docker-based VM infrastructure. We report two metrics: Binary and Partial Accuracy. Binary Accuracy counts a task as successful only when its final score is 1, while Partial Accuracy is the average score over all tasks.

\textbf{Terminal-Bench 2.1}~\cite{terminalbench} is designed to evaluate challenging and realistic command-line tasks. We follow the evaluation setup used for the Qwen-series models, and evaluate on Terminal-Bench 2.1 using Harbor with the Docker backend. Final results report the average score over three independent runs per task, and are compared against the official leaderboard results.

\subsection{Main Results}
\label{sec:exp-results}

\paragraph{Effectiveness on long-horizon computer use.}
Table~\ref{tab:weavebench} reports the results on WeaveBench. The Claude Code harness with Qwen~3.7-Plus achieves a PassRate of 51.8\% and a mean task score of 0.702. Using the same model and retaining Claude Code as the executor backend, LongHorizon-Harness increases these results to 80.7\% and 0.835, respectively. This matched comparison isolates the contribution of the additional task-state management layer. Performance also improves across all eight task domains, showing that the gain is not concentrated in a particular application category. Since the officially reported configurations use a different privilege setting, we include them as reference points but base our main conclusion on the matched Claude Code comparison.

\begin{table}[t]
\centering
\footnotesize
\setlength{\tabcolsep}{4pt}
\renewcommand{\arraystretch}{1.12}
\caption{\textbf{Results on OSWorld~2.0.}
Gray rows denote official results reported by \cite{osworld2} under batched- and single-action settings, while the black row denotes LongHorizon-Harness with Qwen~3.7-Plus in the hybrid setting.
Binary is the percentage of tasks whose final benchmark score equals 1, and Partial is the mean benchmark score over
all 108 tasks. Bold marks the best official result.}
\label{tab:osworld}

\resizebox{0.8\linewidth}{!}{%
\begin{tabular}{l l cc}
\toprule
\textbf{Model} & \textbf{Harness / Mode} & \textbf{Binary}$\uparrow$ & \textbf{Partial}$\uparrow$ \\
\midrule
\g{Claude Opus 4.8} & \g{Batched actions} & \g{\textbf{20.6}} & \g{\textbf{54.8}} \\
\g{Claude Opus 4.7} & \g{Batched actions} & \g{18.2} & \g{48.9} \\
\g{GPT-5.5} & \g{Batched actions} & \g{13.0} & \g{49.5} \\
\midrule
\g{Claude Opus 4.8} & \g{Single action} & \g{18.5} & \g{49.3} \\
\g{Claude Opus 4.7} & \g{Single action} & \g{13.9} & \g{49.1} \\
\g{Claude Sonnet 4.6} & \g{Single action} & \g{8.3} & \g{41.5} \\
\g{MiniMax M3} & \g{Single action} & \g{4.6} & \g{22.3} \\
\g{Kimi 2.6} & \g{Single action} & \g{4.6} & \g{22.1} \\
\g{Qwen 3.7-Plus} & \g{Single action} & \g{2.8} & \g{21.5} \\
\midrule
Qwen 3.7-Plus & LongHorizon-Harness (hybrid) & 8.3 & 35.2 \\
\bottomrule
\end{tabular}%
}
\end{table}

\begin{figure}[t]
\centering
\includegraphics[width=0.9\linewidth]{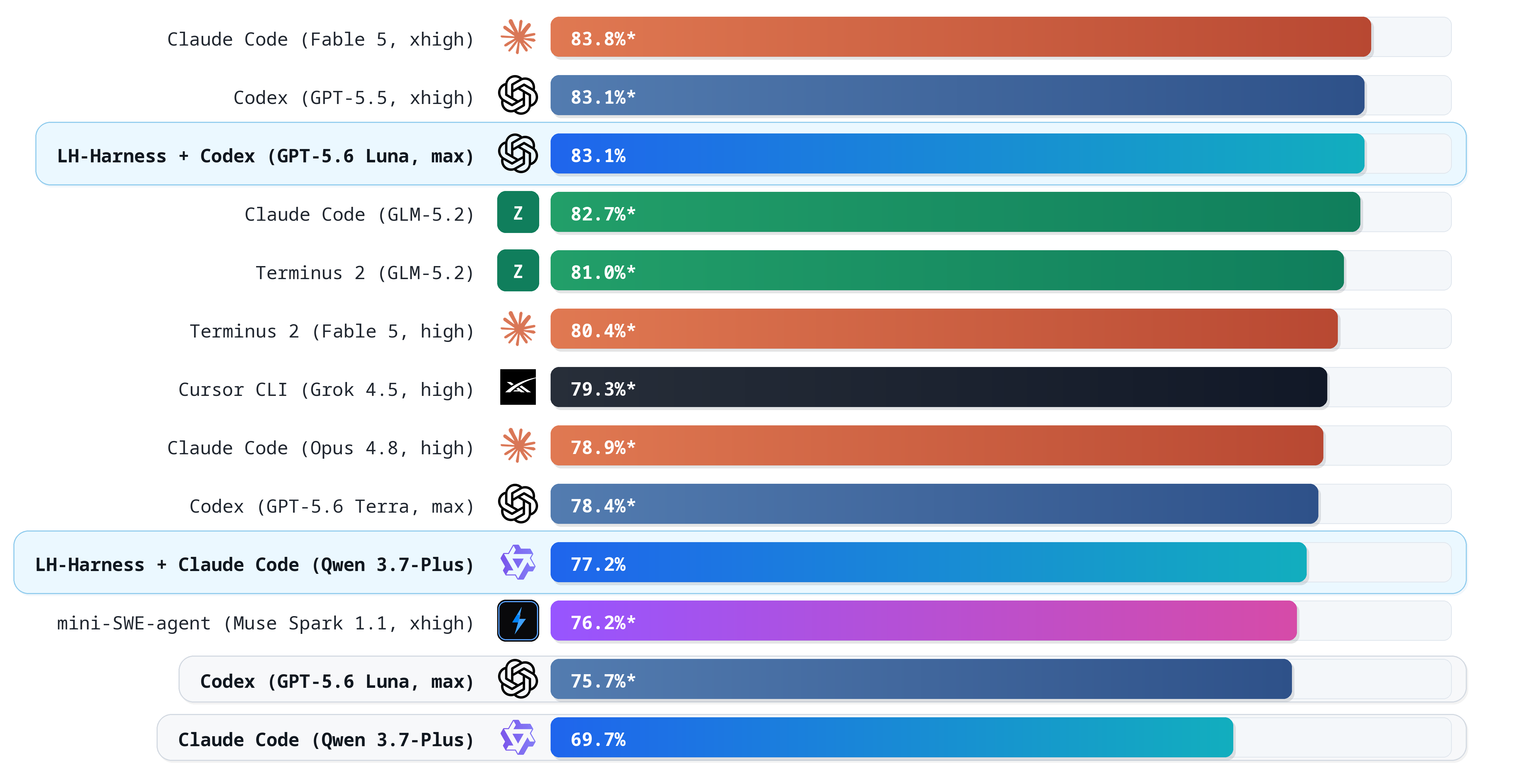}
\caption{\textbf{Results on Terminal-Bench~2.1.}
LongHorizon-Harness retains Claude Code as its executor backend and improves success rate from 69.7\% to 77.2\%. Values marked with $^\ast$ are externally reported metrics.}
\label{fig:tb21}
\end{figure}

\begin{table}[t]
\centering
\footnotesize
\setlength{\tabcolsep}{5pt}
\renewcommand{\arraystretch}{1.15}
\caption{\textbf{OSWorld~2.0 Opus~4.7 subset (34 tasks).}
Both rows use Claude Opus~4.7 as the backbone.}
\label{tab:osworld-opus}

\resizebox{0.8\linewidth}{!}{%
\begin{tabular}{l l cc}
\toprule
\textbf{Model} & \textbf{Harness / Mode} & \textbf{Binary}$\uparrow$ & \textbf{Partial}$\uparrow$ \\
\midrule
Claude Opus 4.7 & \textsc{Single action}          & 20.6          & 55.8          \\
Claude Opus 4.7 & \textbf{LongHorizon-Harness} (hybrid) & \textbf{35.3} & \textbf{66.9} \\
\bottomrule
\end{tabular}%
}
\end{table}

Table~\ref{tab:osworld} reports the results on OSWorld~2.0. With Qwen~3.7-Plus, LongHorizon-Harness increases binary completion from 2.8\% to 8.3\% and partial score from 21.5\% to 35.2\%. The higher partial score shows that the agent satisfies a larger fraction of the task requirements, while the threefold increase in binary completion shows that this progress more frequently leads to a fully completed workflow. Together, the results on WeaveBench and OSWorld~2.0 show that explicit task-state management improves both cross-interface execution and sustained progress over long visual computer-use trajectories.

\paragraph{Complementarity with backbone capability.}
Table~\ref{tab:osworld-opus} further evaluates LongHorizon-Harness with Claude Opus~4.7 on a 34-task subset of OSWorld~2.0. It increases binary completion from 20.6\% to 35.3\% and partial score from 55.8\% to 66.9\%. The consistent improvements with Qwen~3.7-Plus and Claude Opus~4.7 show that the framework does not merely compensate for a weaker model. The backbone determines the quality of the actions and decisions produced within each round, while LongHorizon-Harness determines how reliably verified outcomes are maintained and used across rounds. Stronger models and explicit task-state management therefore provide complementary gains.

\paragraph{Generalization beyond computer use.}
Fig.~\ref{fig:tb21} reports the results on Terminal-Bench~2.1. Due to the limited budget, we only run three configurations ourselves: Claude Code with Qwen~3.7-Plus, LongHorizon-Harness with Claude Code and Qwen~3.7-Plus, and LongHorizon-Harness with Codex and GPT-5.6 Luna. All other results are quoted from the official Terminal-Bench~2.1 leaderboard or the corresponding model's official report.
LongHorizon-Harness improves the success rate of Qwen~3.7-Plus with Claude Code from 69.7\% to 77.2\%. With Codex as the executor backend, LongHorizon-Harness further reaches 83.1\% using GPT-5.6 Luna. Terminal-Bench tasks are performed entirely through the command line and require neither visual perception nor GUI--CLI routing. The gain therefore cannot be attributed to mechanisms specific to computer-use agents. It shows that explicit task-state maintenance, bounded execution contexts, and independent auditing also benefit general long-horizon agent execution. Across hybrid-interface, desktop, and pure CLI environments, LongHorizon-Harness consistently improves how agents preserve and convert intermediate progress into completed tasks.

\subsection{Analysis}
\label{sec:exp-analysis}

We next examine the computational cost of these gains, the task conditions under which they arise, and how the model and harness jointly determine the capability of the resulting agent system.

\begin{figure*}[t]
\centering
\includegraphics[width=1\linewidth]{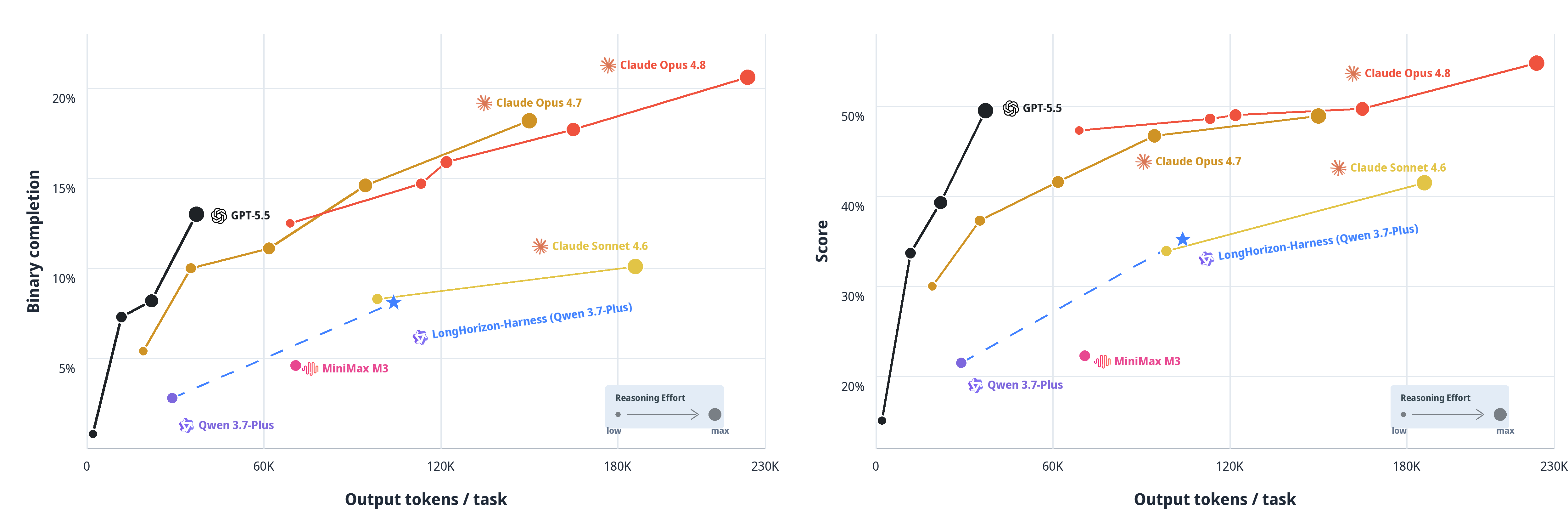}
\caption{\textbf{Cost--performance frontier on OSWorld~2.0.}
Binary completion (left) and partial score (right) versus average output tokens per task.
Colored curves show the official results under different reasoning-effort settings~\cite{osworld2}, with marker size indicating the reasoning effort.
The blue star denotes LongHorizon-Harness with Qwen~3.7-Plus, and the dashed arrow shows its improvement over the official single-action Qwen~3.7-Plus baseline.}
\label{fig:osworld-frontier}
\end{figure*}

\paragraph{Cost--performance trade-off.}
Fig.~\ref{fig:osworld-frontier} shows that LongHorizon-Harness moves Qwen~3.7-Plus to a substantially stronger point on the OSWorld~2.0 cost--performance frontier. The same model improves from 2.8\% to 8.3\% in binary completion and from 21.5\% to 35.2\% in partial score, although its average output-token consumption increases from 28.9K to 104K per task. This result shows that a stronger harness can raise the task-level performance achievable by a fixed model. The increased cost observed with Qwen~3.7-Plus is specific to this model--task configuration; as analyzed below, the token cost of LongHorizon-Harness depends strongly on the capability of the underlying executor.

\begin{figure*}[t]
\centering
\includegraphics[width=\linewidth]{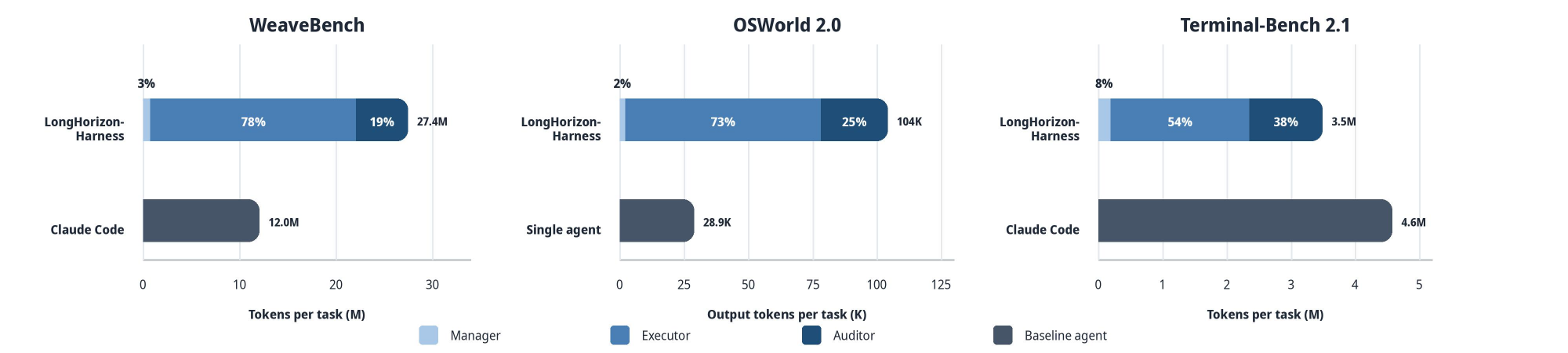}
\caption{\textbf{Token consumption by role.}
Average tokens per task consumed by the manager, executor, and auditor in LongHorizon-Harness, compared with the corresponding baseline.
Segment labels report the fraction consumed by each role.
All LH-Harness roles use Qwen~3.7-Plus.
WeaveBench and Terminal-Bench~2.1 report total tokens, whereas OSWorld~2.0 reports output tokens because the official provides only output-token statistics~\cite{osworld2}.}
\label{fig:role-tokens}
\end{figure*}

\paragraph{Computation across roles.}
Fig.~\ref{fig:role-tokens} decomposes the token consumption of LongHorizon-Harness across the manager, executor, and auditor. The manager accounts for only 2.8\%, 2.0\%, and 8.1\% of the total tokens on WeaveBench, OSWorld~2.0, and Terminal-Bench~2.1, respectively, showing that explicit task-state maintenance introduces little computational overhead. The auditor accounts for 19.4\%, 24.8\%, and 38.1\% of the tokens, indicating that independent verification constitutes the main additional investment of the framework. The overall token change nevertheless differs across benchmarks. LongHorizon-Harness consumes $2.3\times$ the baseline tokens on WeaveBench and $3.6\times$ the baseline output tokens on OSWorld~2.0, whereas it consumes 24\% fewer tokens on Terminal-Bench~2.1 while achieving a higher success rate. These results show that the framework does not impose a fixed token multiplier. Its total cost depends on both the task and the number of execution and recovery rounds required by the underlying model.

\begin{figure*}[ht]
\centering
\includegraphics[width=\linewidth]{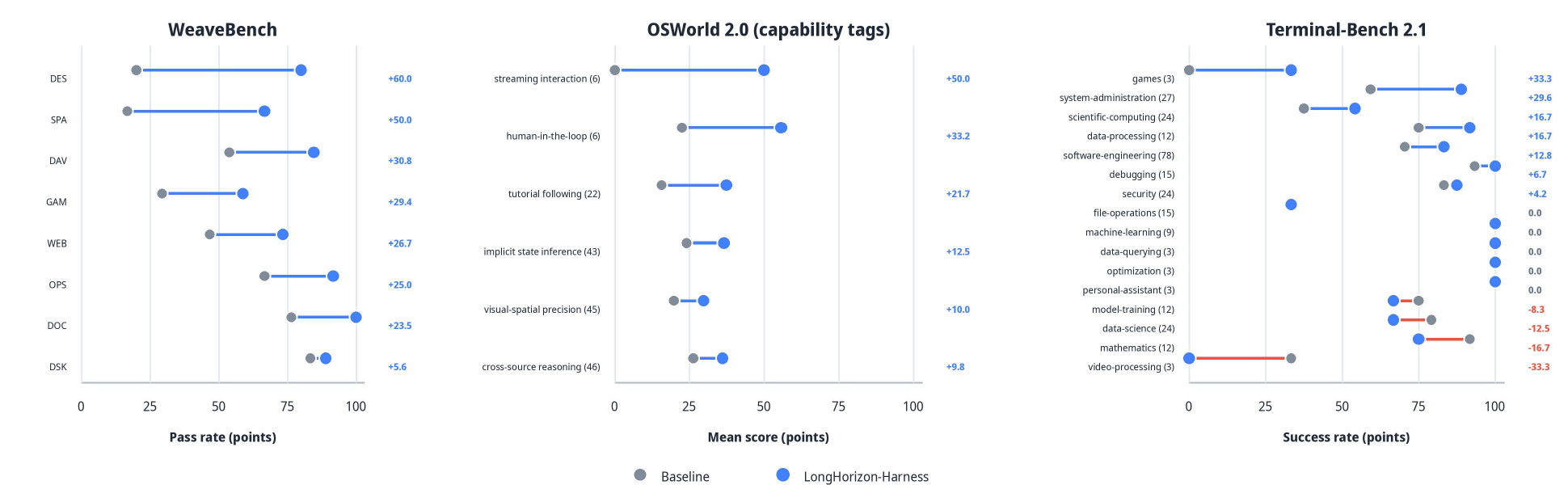}
\caption{\textbf{Performance gains across task types.}
LongHorizon-Harness (blue) is compared with the same-model baseline (gray) across WeaveBench domains (left), OSWorld~2.0 capability tags (middle), and Terminal-Bench~2.1 categories (right).
The rightmost column of each panel reports the absolute performance change in points.
Blue connectors indicate improvements, and red connectors indicate regressions.
Parenthesized values report the number of OSWorld tasks or Terminal-Bench trajectories in each category.}
\label{fig:domain-deltas}
\end{figure*}

\begin{table*}[h]
\centering
\footnotesize
\setlength{\tabcolsep}{5pt}
\renewcommand{\arraystretch}{1.12}
\caption{\textbf{Per-task scores and token consumption on the WeaveBench Games subset} (17 tasks).
CC = Claude Code; LH = LongHorizon-Harness wrapping Claude Code as its backend.
Tok denotes total tokens per task (M).
\textbf{Bold} marks the better score per backbone.}
\label{tab:games}
\begin{tabular}{l cc cc | cc cc}
\toprule
& \multicolumn{4}{c}{\textbf{Claude Opus 4.7}} & \multicolumn{4}{c}{\textbf{Qwen 3.7-Plus}} \\
\cmidrule(lr){2-5} \cmidrule(lr){6-9}
& \multicolumn{2}{c}{Score} & \multicolumn{2}{c}{Tok (M)} & \multicolumn{2}{c}{Score} & \multicolumn{2}{c}{Tok (M)} \\
\cmidrule(lr){2-3} \cmidrule(lr){4-5} \cmidrule(lr){6-7} \cmidrule(lr){8-9}
\textbf{Task} & CC & LH & CC & LH & CC & LH & CC & LH \\
\midrule
\texttt{mines\_solve}           & \textbf{0.958} & 0.920 &  8.8 &  4.9 & 0.040 & \textbf{0.590} &  6.0 & 97.2 \\
\texttt{game\_ui\_bug}          & 0.830 & \textbf{0.910} & 11.7 &  8.3 & 0.750 & \textbf{0.850} & 10.0 & 34.2 \\
\texttt{stockfish\_puzzle}      & \textbf{0.603} & 0.550 &  7.5 &  3.4 & 0.000 & \textbf{0.550} &  0.9 & 15.2 \\
\texttt{gdb\_pygame\_cheat}     & 0.920 & \textbf{0.950} & 33.9 &  5.6 & 0.720 & \textbf{0.860} &  5.9 & 13.9 \\
\texttt{mines\_visual}          & 0.800 & \textbf{0.860} & 29.1 &  4.3 & 0.000 & \textbf{0.470} &  3.7 & 50.1 \\
\texttt{quadrapassel\_autoplay} & 0.600 & \textbf{0.870} & 50.5 & 13.4 & 0.000 & \textbf{0.300} & 10.0 & 59.1 \\
\texttt{pokerth\_equity\_play}  & 0.050 & \textbf{0.400} &  4.6 & 23.1 & 0.000 & \textbf{0.920} &  9.0 & 16.2 \\
\texttt{sokoban\_solver}        & \textbf{0.990} & 0.950 &  4.4 &  3.3 & \textbf{0.960} & 0.958 &  3.9 &  4.2 \\
\texttt{rhythm\_autoplay}       & 0.880 & \textbf{0.900} & 10.5 &  6.9 & \textbf{0.950} & 0.830 &  8.5 & 19.6 \\
\texttt{gnuchess\_blunder\_hunt}& 0.720 & \textbf{0.920} & 20.1 & 14.0 & 0.830 & \textbf{0.940} & 19.5 & 21.4 \\
\texttt{anagramarama\_word\_grid}& 0.550 & \textbf{0.750} &  6.6 & 18.3 & 0.750 & \textbf{0.830} & 12.8 & 12.5 \\
\texttt{godot\_scene\_debug}    & \textbf{0.958} & 0.920 & 11.4 &  7.8 & \textbf{0.930} & 0.870 &  8.6 & 17.1 \\
\texttt{hedgewars\_mission\_debug}& 0.790 & \textbf{0.810} & 19.6 &  8.5 & 0.740 & \textbf{0.830} & 14.4 & 13.1 \\
\texttt{supertux\_level\_repair}& 0.600 & \textbf{0.720} & 19.7 & 11.8 & 0.000 & \textbf{0.650} &  7.5 & 21.0 \\
\texttt{xmoto\_level\_repair}   & \textbf{0.600} & 0.550 &  8.1 & 23.4 & \textbf{0.920} & 0.850 & 19.4 & 83.3 \\
\texttt{kmahjongg\_pair\_solver}& 0.350 & \textbf{0.830} &  8.8 & 13.2 & \textbf{0.790} & 0.750 & 30.1 & 35.4 \\
\texttt{pysol\_freecell}        & 0.360 & \textbf{0.940} & 24.8 & 18.0 & \textbf{0.520} & 0.410 & 12.5 & 68.8 \\
\midrule
\textbf{Mean}                   & 0.680 & \textbf{0.809} & 16.5 & 11.1 & 0.524 & \textbf{0.733} & 10.7 & 34.3 \\
\bottomrule
\end{tabular}
\end{table*}

\paragraph{Task-dependent effectiveness.}
Fig.~\ref{fig:domain-deltas} compares the gains of LongHorizon-Harness across task types. On WeaveBench, Design and Spatial/3D obtain the largest PassRate improvements of 60.0 and 50.0 points, respectively, whereas the Desktop domain, which already reaches 83.3\% with the baseline, improves by 5.6 points. On OSWorld~2.0, some of the largest gains appear in streaming interaction, human-in-the-loop, and tutorial-following tasks. On Terminal-Bench~2.1, system-administration and game tasks improve substantially, while several short analytical categories show smaller gains or regressions. Despite operating through different interfaces, the categories with larger improvements commonly require agents to preserve, inspect, and revise multiple dependent environment states over an extended trajectory. The smaller improvements on several analytical categories suggest that LongHorizon-Harness provides less benefit when task performance is dominated by an individual model capability, such as visual perception, mathematical reasoning, coding, or algorithm design. Independent auditing can detect an incorrect result and initiate recovery, but it cannot supply a capability that the model does not possess. The effectiveness of LongHorizon-Harness is thus determined less by whether a task uses a GUI or CLI than by whether its primary bottleneck lies in long-horizon execution reliability or in solving an individual step.

\paragraph{Agent capability as a system property.}
Table~\ref{tab:games} compares the Claude Code baseline and LongHorizon-Harness on the same 17 WeaveBench Games tasks using Claude Opus~4.7 and Qwen~3.7-Plus. LongHorizon-Harness improves the mean score of Opus from 0.680 to 0.809 and that of Qwen from 0.524 to 0.733, with the largest gains concentrated on tasks where the corresponding baseline performs poorly. All six tasks on which the Qwen baseline scores at or below 0.04 recover to scores between 0.30 and 0.92, showing that the framework primarily raises the failure floor by recovering trajectories that would otherwise end in near-total failure. Qwen with LongHorizon-Harness further reaches a mean score of 0.733, exceeding the 0.680 obtained by Opus with the base Claude Code harness. This comparison demonstrates that agent capability is a property of the complete model--harness system. The model determines the actions and solutions available within each execution round, while the harness determines how reliably these local capabilities are decomposed, verified, recovered, and accumulated into end-to-end task completion. A stronger harness can consequently raise the task-level performance achievable with a fixed model, even though it does not add new primitive capabilities to that model. The token results reveal the same dependence on both components: LongHorizon-Harness increases Qwen's average consumption from 10.7M to 34.3M tokens per task, but reduces Opus consumption from 16.5M to 11.1M. This opposite pattern suggests that a stronger model can satisfy subtask contracts with fewer audit--replan rounds, whereas a weaker model must spend additional inference on repeated execution and recovery. Model capability and harness design jointly determine both the achievable performance and the computational cost of a long-horizon agent.

\subsection{Case Studies}
\label{sec:case-studies}

We further compare paired trajectories in which the Claude Code
baseline and LongHorizon-Harness execute the same task using the same
Qwen~3.7-Plus model. These cases examine what persists between rounds
under the Manage-Execute-Audit loop. Unresolved failures and completed
progress are carried outside the execution history, completion claims
are checked before entering the task state, and unmet dependencies
remain available when the manager defines subsequent subtasks. The
examples also illustrate how persistent task state connects executors
that operate in separate, fresh contexts.

\paragraph{Recovering from a stalled interaction.}
Fig.~\ref{fig:case-webrtc} compares how the two systems proceed after
the same GUI interaction fails. The baseline recognizes that the
Wireshark ``Decode As'' dialog is unresponsive, but this observation
remains embedded in its growing execution history. It consequently
returns to the same interaction for more than 400 steps without
collecting the remaining task evidence. LongHorizon-Harness instead
records the failed interaction and the unresolved evidence gaps in the
task state. Later executors focus directly on the missing
simulcast-layer charts and packet-level evidence, increasing the score
from 0.59 to 0.92. By externalizing the stalled interaction,
LongHorizon-Harness preserves the progress already verified and
provides the next executor with only the unresolved requirements.
Recovery therefore resumes from the latest audited task state rather
than from the failed trajectory that produced it.

\paragraph{Auditing apparent completion.}
Fig.~\ref{fig:case-heading} presents a case in which the executor stops
at a visually plausible result that does not satisfy the full task
specification. The baseline directly edits the document XML and
terminates after the headings appear correct. However, the task
requires the styles to be applied through the LibreOffice workflow,
and the result therefore receives a score of 0.00.
LongHorizon-Harness performs the prescribed GUI operations, confirms
the style changes during execution, and rechecks the document page by
page. The auditor then parses the document XML and confirms that all
15 headings have the required underlying style, yielding a score of
0.89. This independent audit prevents the executor's plausible but
non-compliant completion claim from becoming part of the persistent
task state. Subsequent planning can instead proceed from the document
state confirmed in the environment, rather than from the executor's
assessment of its own result.

\paragraph{Preserving pre-repair evidence.}
Fig.~\ref{fig:case-vlookup} presents a task in which part of the
required evidence belongs to the environment state before the repair.
The baseline captures an initial view of the formula errors, but
modifies the spreadsheet before completing the full pre-repair evidence
sequence. Its direct XML edit also leaves formula errors after the file
is reopened, producing an inconsistent before-and-after record.
LongHorizon-Harness records the missing pre-repair evidence as a
pending requirement. The next executor first completes this evidence,
then repairs the spreadsheet and checks the resulting formula cells.
The auditor subsequently reviews the complete sequence of nine
screenshots, increasing the score from 0.45 to 0.87. Keeping the
pre-repair evidence explicit changes the next subtask selected from the
task state: the original spreadsheet state is documented before the
repair is allowed to modify it. The resulting execution order preserves
a consistent evidence chain across the state transition.

\paragraph{Continuing from verified progress.}
Fig.~\ref{fig:case-lighthouse} compares two trajectories in which the
model can perform the central optimization but differs in how the
remaining workflow is maintained. The baseline improves the target
page, yet the same growing session must continue operating DevTools,
tracking the remaining deliverables, collecting evidence, and assessing
its own progress. Repeated DevTools interactions eventually exhaust
the available budget, leaving the task incomplete.
LongHorizon-Harness retains the completed optimization in the task
state and derives subsequent subtasks from the remaining requirements.
Fresh-context executors then confirm the page state, inspect the
relevant script structure, and collect the performance trace, while
the auditor checks the final screenshot against its metadata. The
score increases from 0.53 to 0.85. Because the completed optimization
persists outside the execution history, later executors can focus on
the remaining evidence without reconstructing or carrying the preceding
optimization process. The manager maintains continuity across the full
workflow, while each executor uses its context for the current
environment transition, allowing the model's local progress to be
carried forward into a more complete task result.

\DeclareRobustCommand{\taskid}[1]{\texttt{\detokenize{#1}}}

\begin{figure*}[h]
\centering
\includegraphics[width=\linewidth]{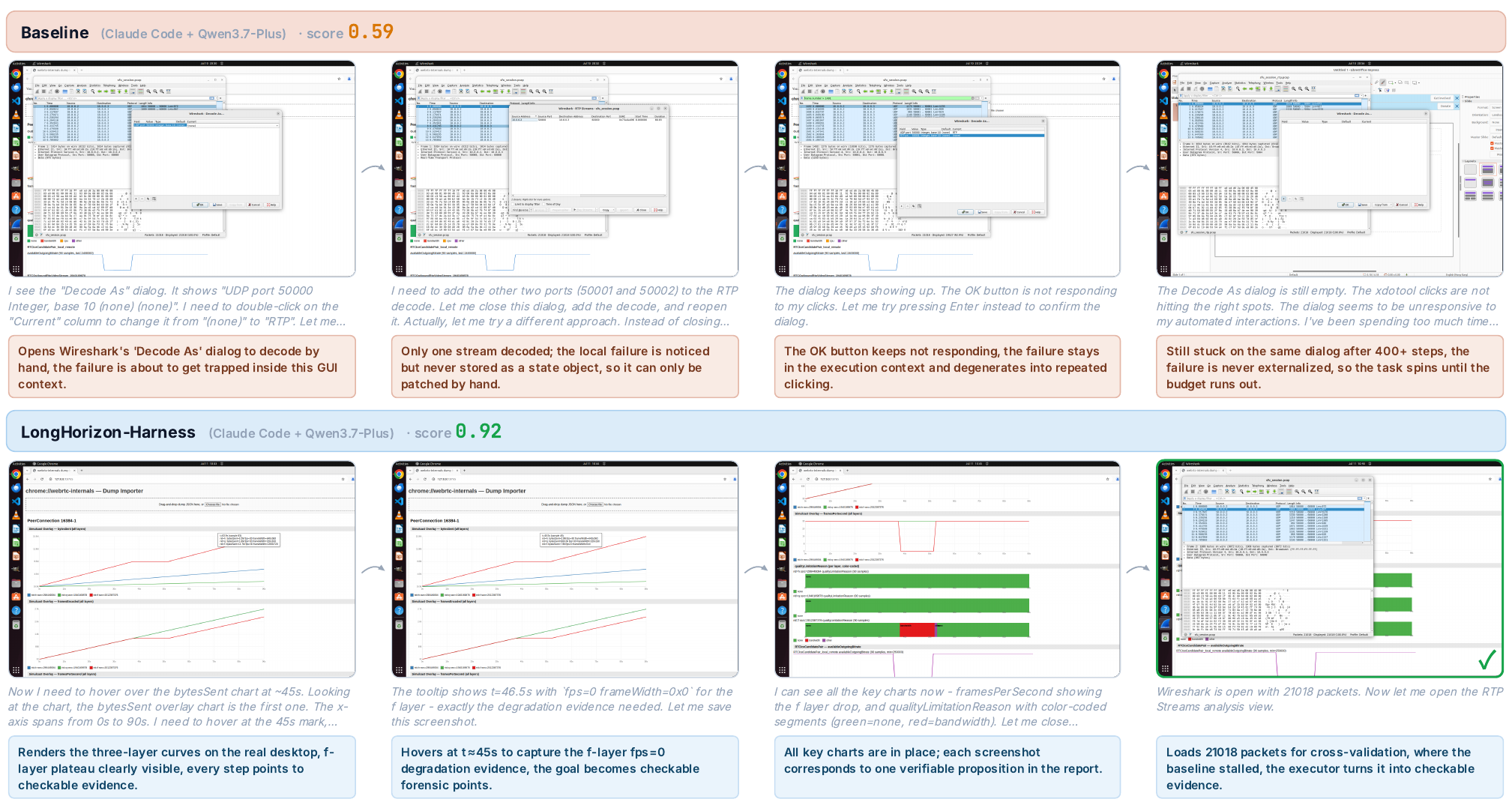}
\caption[Recovering from a stalled interaction]{
\textbf{Recovering from a stalled interaction.}
The case is \taskid{WEB_task_16}, a WebRTC simulcast-layer audit.
\emph{Top:} The Claude Code baseline recognizes that Wireshark's
``Decode As'' dialog is unresponsive, but continues retrying the same
interaction for more than 400 steps and obtains a score of 0.59.
\emph{Bottom:} LongHorizon-Harness records the unresolved evidence gaps
in the task state. Subsequent execution rounds collect the missing
chart- and packet-level evidence, resulting in a score of 0.92.}
\label{fig:case-webrtc}
\end{figure*}

\begin{figure}[H]
\centering
\includegraphics[width=\linewidth]{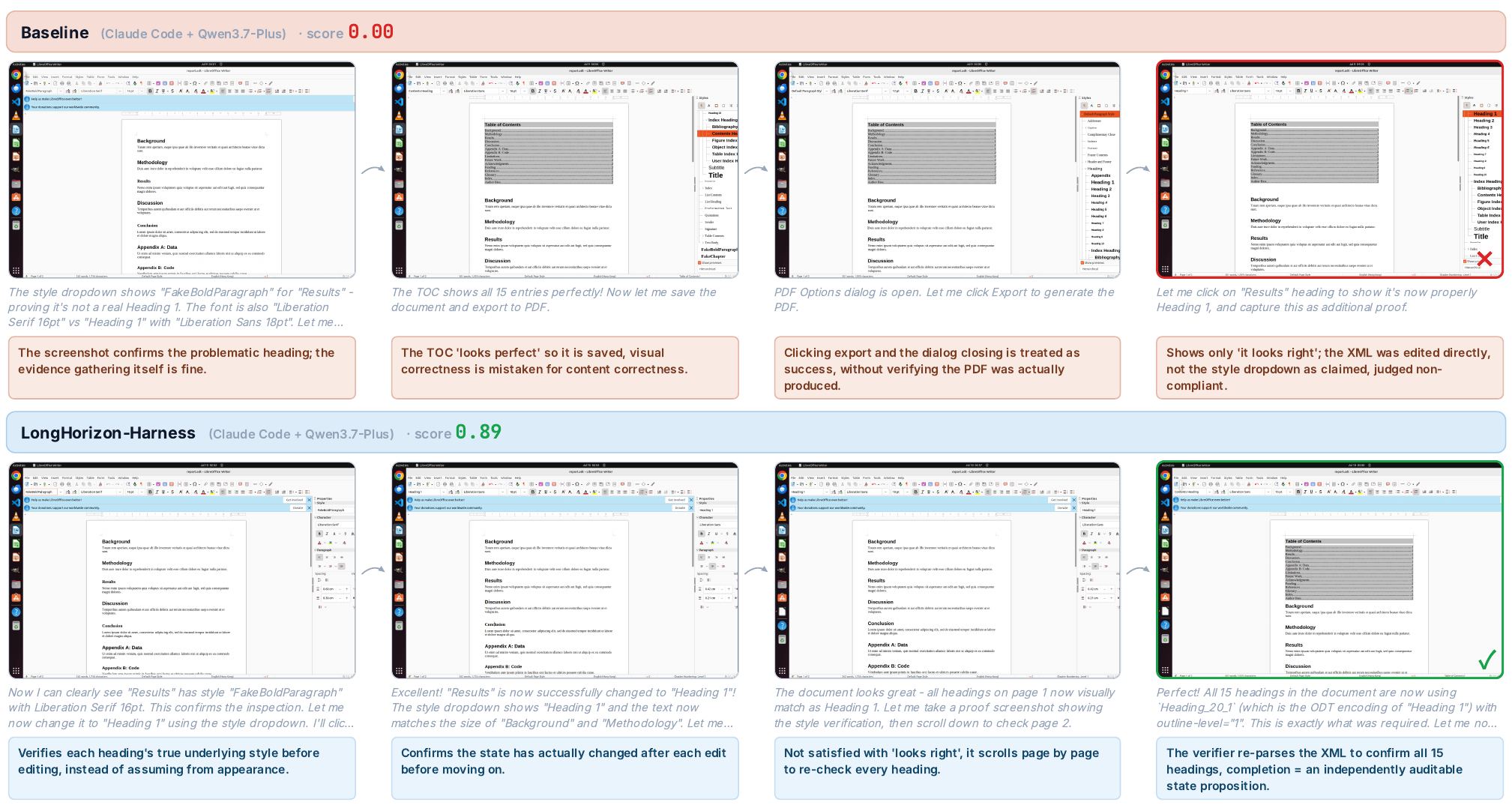}
\caption[Auditing apparent completion]{
\textbf{Auditing apparent completion.}
The case is \taskid{DOC_task_2}, which requires heading-style
normalization.
\emph{Top:} The Claude Code baseline edits the document XML directly
and terminates with a visually plausible result. Because the required
LibreOffice workflow is not followed, the result receives a score of
0.00.
\emph{Bottom:} LongHorizon-Harness applies the heading styles through
the prescribed GUI workflow and rechecks the document. The auditor then
parses the document XML to confirm the final style of all 15 headings,
obtaining a score of 0.89.}
\label{fig:case-heading}
\end{figure}

\begin{figure}[h]
\centering
\includegraphics[width=\linewidth]{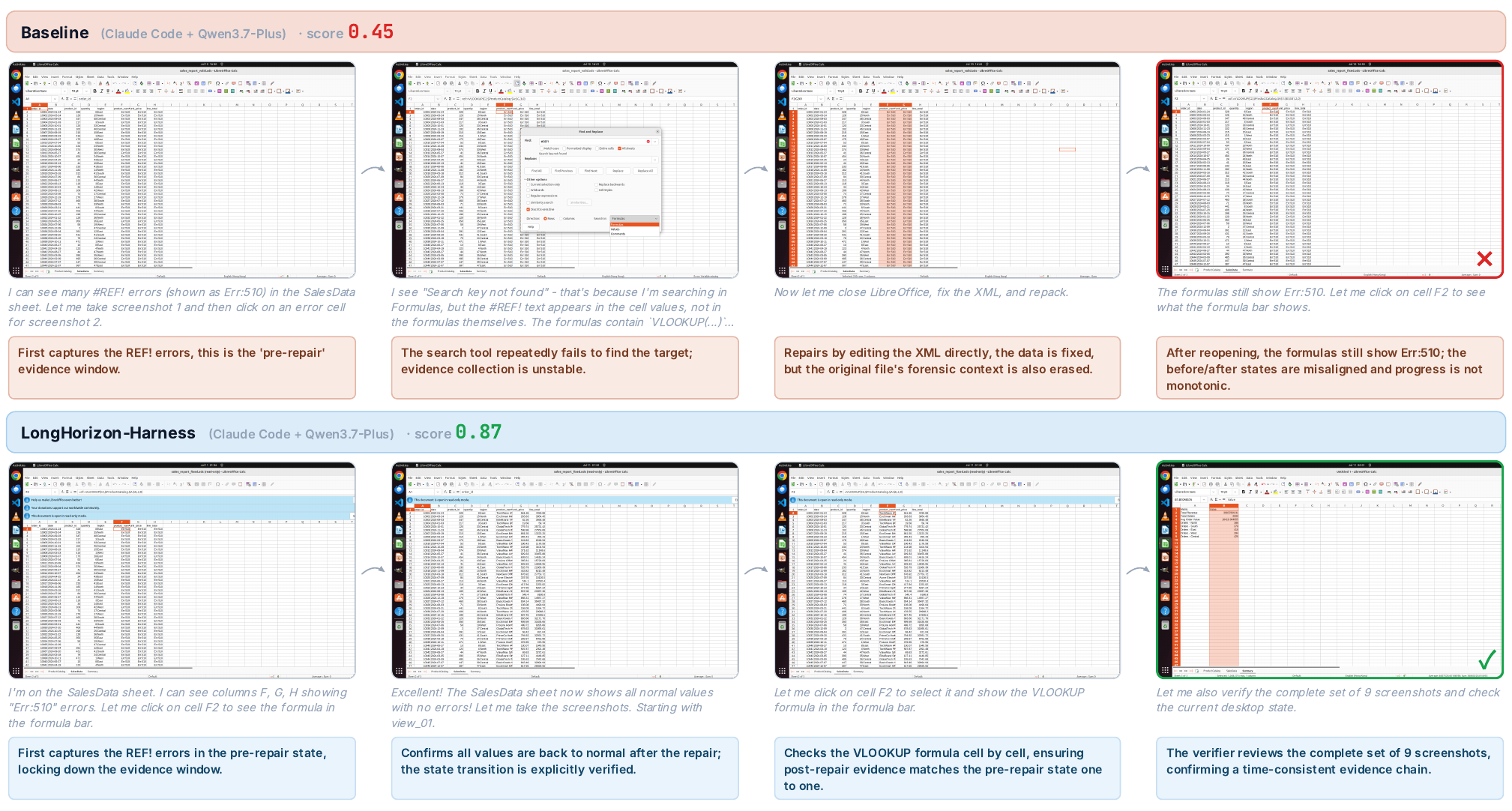}

\caption[Preserving pre-repair evidence]{
\textbf{Preserving pre-repair evidence.}
In \taskid{DOC_task_4}, the agent must repair a Calc VLOOKUP formula while preserving before-and-after evidence.
\emph{Top:} The Claude Code baseline records initial formula errors but edits the spreadsheet before completing the required pre-repair evidence, leaving the final file and evidence sequence inconsistent and scoring 0.45.
\emph{Bottom:} LongHorizon-Harness keeps the missing pre-repair evidence as a pending requirement, records it before modification, and audits the full nine-screenshot sequence, scoring 0.87.}
\label{fig:case-vlookup}
\end{figure}

\begin{figure}[H]
\centering
\includegraphics[width=\linewidth]{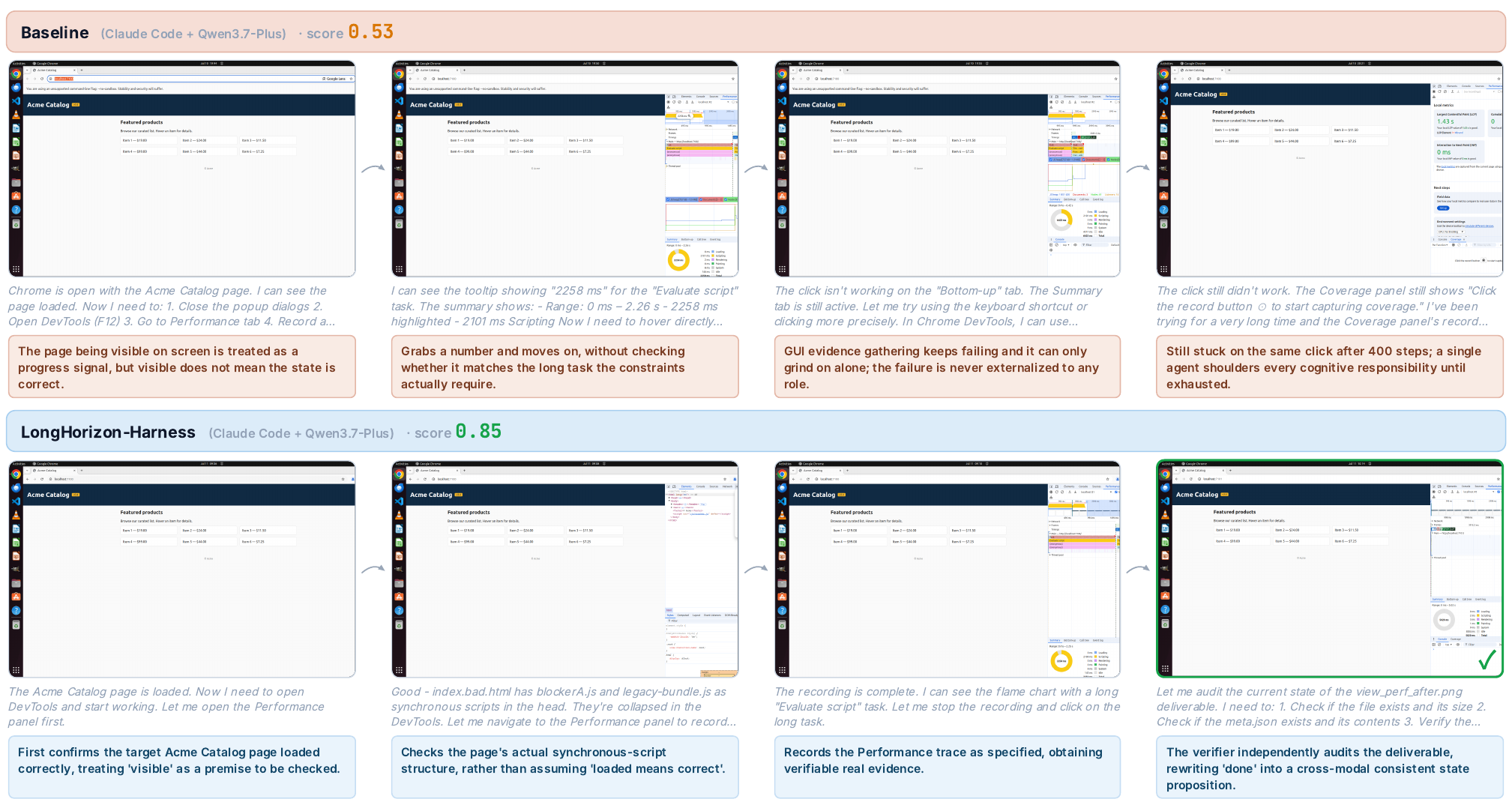}

\caption[Continuing from verified progress]{
\textbf{Continuing from verified progress.}
\taskid{WEB_task_10} is a Lighthouse performance-optimization task.
\emph{Top:} The Claude Code baseline completes the core optimization but gets stuck during DevTools evidence collection, fails to produce all required deliverables, and scores 0.53.
\emph{Bottom:} LongHorizon-Harness preserves the verified optimization state, delegates the remaining evidence requirements to later rounds, and audits the final screenshot and metadata, scoring 0.85.}
\label{fig:case-lighthouse}
\end{figure}

%% file: sec/appendix.tex
\appendix
\section{Detailed Experimental Setup}
\subsection{WeaveBench}
WeaveBench~\cite{weavebench} consists of 114 long-horizon computer-use tasks spanning eight domains: Desktop Applications, Document Processing, Games, Web Development, Data Analysis and Visualization, DevOps, Spatial/3D Applications, and Design. Each task requires the agent to use both GUI and CLI interactions within the same workflow, rather than relying solely on command-line or browser operations. We follow the standard WeaveBench evaluation protocol: each task is launched in an isolated containerized Ubuntu desktop VM, and task executions do not share runtime state. The task definitions, workspaces, runtime assets, judge templates, and scoring procedure are all taken from the official release. The evaluator uses Claude Opus 4.7 as a trajectory-aware judge. We report two aggregate metrics: PassRate, defined as the fraction of tasks with score at least 0.8, and Overall, defined as the average score over all 114 tasks. In addition to the aggregate results, we report domain-level PassRate for the eight domains: Desktop Applications (DSK), Document Processing (DOC), Games (GAM), Web Development (WEB), Data Analysis and Visualization (DAV), DevOps (OPS), Spatial/3D Applications (SPA), and Design (DES).

For the model and execution framework, we evaluate Qwen 3.7-Plus. The Claude Code runtime version is \texttt{claude-code 2.1.76}. Each task is allowed up to 25 Manage-Execute-Audit rounds. The per-round timeout is 1800 seconds for the executor, and 300 seconds for both the manager and the verifier. The Claude Code reasoning effort on the task side is set to \texttt{high}; the judge is executed with OpenClaw, with \texttt{AJ\_THINKING=medium}.

For GUI tools, we preserve the basic WeaveBench computer-use interface, including the \texttt{screenshot} perception tool and desktop actuation tools implemented with \texttt{pyautogui}, such as click, double-click, mouse move, keyboard input, scroll, drag, and wait. In addition, we introduce a restricted evidence-preservation tool, \texttt{save\_screenshot}. This tool does not provide any new environment-control capability, nor does it expose additional information to the agent; it can only save the currently visible real desktop screen to the target directory as an official screenshot artifact required by certain tasks.

Since GUI tasks frequently provide screenshot inputs, we use an image URL proxy for visual inputs: screenshots are uploaded to an image-hosting backend and referenced as image URLs in model requests, avoiding oversized request bodies caused by multiple base64-encoded screenshots.

\subsection{OSWorld 2.0}
OSWorld 2.0~\cite{osworld2} consists of 108 professional desktop workflow tasks, with a median human completion time of approximately 1.6 hours. We use the official OSWorld-v2 \texttt{osworld-v2-2026.06.24} release, including the corresponding task definitions, task assets, Docker VM image, and mocked websites. Experiments are conducted using the standard OSWorld Docker-based VM infrastructure in an Ubuntu desktop environment, with a default resolution of $1920\times1080$. Final scoring is performed by the native OSWorld \texttt{env.evaluate()} procedure.

Unlike the main official baselines, which typically rely only on GUI actions, LongHorizon-Harness uses a hybrid tool pool on OSWorld-v2. GUI actions are executed by a computer MCP server inside the VM, supporting desktop operations such as clicking, typing, dragging, and hotkeys. The CLI role can execute shell commands in the same VM for filesystem inspection, script execution, and long-context task processing. These tools are hosted inside the VM through a fixed version of \texttt{claude-code 2.1.176}.

For model inference, we set \texttt{thinking.type=enabled}, \texttt{max\_tokens}=65{,}536, and \texttt{effort=max}. The native OSWorld evaluator and user simulator use the same \texttt{qwen3.7-plus} configuration as the execution agent. For human-in-the-loop tasks in the benchmark that require user information or authorization, we do not introduce manual intervention; instead, we use the native OSWorld user-simulator interface to provide responses.

We report two metrics: Binary Accuracy and Partial Accuracy. Binary Accuracy counts a task as successful only when its final score is 1, while Partial Accuracy is the average fine-grained score over all 108 tasks and measures partial task completion.

\subsection{Terminal-Bench 2.1}
Terminal-Bench 2.1~\cite{terminalbench} is designed to evaluate agents on challenging and realistic software-engineering and command-line tasks. We use Harbor as the evaluation framework, with the Docker backend providing an isolated execution environment for each task. During evaluation, we preserve the CPU, memory, and environment constraints originally defined for each task.

To ensure consistency with existing evaluations of Qwen-series models, both the baseline and CUA-Harness settings use Claude Code as the underlying CLI execution agent, connected to the \texttt{qwen3.7-plus} model. The Claude Code version is \texttt{v2.1.211}. Since the Claude Code command-line interface does not directly expose all low-level request parameters, we introduce a transparent request proxy inside the container. All model requests are configured with \texttt{temperature}=1.0, \texttt{top\_p}=0.95, \texttt{top\_k}=20, \texttt{enable\_thinking=true}, and \texttt{max\_tokens}=65{,}536.

Each trial has a timeout of 5 hours. We run three independent trials per task, report the average score over the three runs for each task, and compare against the Claude Code baseline under the same model setting as well as the official leaderboard results.

\section{Detailed Experimental Results}
\subsection{Detailed Results on the OSWorld 2.0 Opus 4.7 Subset}

Table~\ref{tab:osworld-opus47-detailed} reports the per-task scores on the 34-task OSWorld 2.0 Opus 4.7 subset. The baseline uses the standard single-action GUI setting, while LongHorizon-Harness uses our hybrid GUI+CLI tool pool. On this subset, LongHorizon-Harness improves the partial score from $55.83\%$ to $66.86\%$, and the binary accuracy from $20.6\%$ to $35.3\%$.

\begin{table*}[t]
\centering
\caption{Detailed per-task results on the OSWorld 2.0 Opus 4.7 subset. Ours denotes LongHorizon-Harness.}
\setlength{\tabcolsep}{7pt}
\begin{tabular}{lrrlrr}
\toprule
Task & Baseline & Ours & Task & Baseline & Ours \\
\midrule
001 & 0.7778 & 1.0000 & 018 & 0.0000 & 0.0000 \\
002 & 0.7500 & 1.0000 & 019 & 0.5200 & 0.7600 \\
003 & 1.0000 & 1.0000 & 020 & 0.1000 & 1.0000 \\
004 & 0.8000 & 0.7667 & 021 & 1.0000 & 0.0000 \\
005 & 0.0000 & 1.0000 & 022 & 1.0000 & 1.0000 \\
006 & 0.5556 & 0.8889 & 023 & 0.9300 & 0.9300 \\
007 & 0.4000 & 0.8000 & 025 & 0.2401 & 0.0833 \\
010 & 1.0000 & 1.0000 & 027 & 0.5000 & 0.0000 \\
011 & 0.2963 & 0.5556 & 028 & 0.4000 & 1.0000 \\
012 & 0.6000 & 1.0000 & 030 & 1.0000 & 0.0000 \\
013 & 1.0000 & 0.7500 & 032 & 0.9111 & 0.2000 \\
014 & 0.9000 & 0.7400 & 033 & 0.1000 & 1.0000 \\
015 & 0.2840 & 0.7667 & 034 & 0.5938 & 0.8750 \\
016 & 0.8571 & 0.7524 & 051 & 0.0000 & 0.6000 \\
017 & 0.1429 & 0.0000 & 052 & 1.0000 & 1.0000 \\
060 & 0.1000 & 0.1000 & 054 & 0.3628 & 1.0000 \\
061 & 0.3574 & 0.3891 & 062 & 0.5033 & 0.7753 \\
\bottomrule
\end{tabular}

\label{tab:osworld-opus47-detailed}
\end{table*}

\subsection{Fine-Grained Breakdown by Benchmark and Task Type}
\label{app:fine-grained-breakdown}

Beyond the Opus 4.7 subset reported above, we provide a fine-grained breakdown of the main evaluation results by benchmark, domain, task property, and failure mode. We use LongHorizon-Harness (LH-Harness) as the abbreviated name in the tables. The purpose of this analysis is to identify where the harness changes the execution dynamics, rather than only reporting aggregate gains. Across benchmarks, the largest improvements appear when progress can be represented as verifiable environment state: installed binaries, repository states, screenshots with well-defined metadata, application configuration, structured files, or cross-checked evidence from multiple interfaces. The remaining failures concentrate in tasks where the decisive condition is difficult to close under the available verifier, such as hidden performance thresholds, embodied visual precision, temporal video evidence, or task semantics that admit several plausible interpretations.

\subsubsection{WeaveBench}
\label{app:fine-grained-weavebench}

Table~\ref{tab:weavebench-domain-breakdown} reports the domain-level breakdown on WeaveBench. LH-Harness improves the mean score in seven of the eight domains, with the largest gains in Design (DES), Spatial/3D Applications (SPA), and Games (GAM). Pass rate also increases in every domain, including Desktop (DSK), where the mean score slightly decreases despite a higher pass rate. This pattern indicates that the harness mainly raises the lower tail of long-horizon trajectories: it converts many near-failures into partial or complete successes, while a small number of already-strong baseline trajectories can be affected by additional verification and repair steps.

\begin{table}[t]
\caption{WeaveBench domain-level results. Score $\Delta$ is LH-Harness minus baseline. Pass-rate improvement is reported in percentage points.}
\label{tab:weavebench-domain-breakdown}
\centering
\small
\setlength{\tabcolsep}{5pt}
\begin{tabular}{lrrrr}
\toprule
Domain & Baseline & LH-Harness & Score $\Delta$ & Pass-rate $\Delta$ \\
\midrule
DES & 0.5630 & 0.8160 & +0.2530 & +60.00 pp \\
SPA & 0.5800 & 0.8192 & +0.2392 & +50.00 pp \\
GAM & 0.5235 & 0.7328 & +0.2093 & +29.41 pp \\
DAV & 0.6981 & 0.8608 & +0.1627 & +30.77 pp \\
OPS & 0.7758 & 0.9090 & +0.1332 & +25.00 pp \\
DOC & 0.8009 & 0.8929 & +0.0920 & +23.53 pp \\
WEB & 0.7287 & 0.8147 & +0.0860 & +26.67 pp \\
DSK & 0.8671 & 0.8465 & -0.0205 & +5.56 pp \\
\bottomrule
\end{tabular}
\end{table}

The Games domain provides a useful stress test because it combines visual interaction, symbolic state, search, and environment-specific constraints. Table~\ref{tab:weavebench-game-breakdown} shows that the largest gains occur on tasks where the baseline often fails to establish a stable state representation. For example, Mines, Stockfish puzzle analysis, Quadrapassel autoplay, PokerTH equity play, and SuperTux level repair all improve from near-zero or zero baseline scores to nontrivial completion. In these cases, the harness helps by decomposing the task into explicit subgoals and by carrying forward verified facts rather than a long interaction transcript. In contrast, when the baseline already solves the core task, additional rounds provide less benefit and can even slightly reduce score, as in rhythm autoplay, X-Moto, KMahjongg, and FreeCell.

\begin{table*}[t]
\caption{Fine-grained results for WeaveBench Games tasks. Task names omit the shared \texttt{GAM\_task\_*} prefix for compactness.}
\label{tab:weavebench-game-breakdown}
\centering
\scriptsize
\setlength{\tabcolsep}{4pt}
\begin{tabular}{lrr@{\hspace{1.2em}}lrr}
\toprule
Task & Baseline & LH-Harness & Task & Baseline & LH-Harness \\
\midrule
gnome\_mines\_solve & 0.040 & 0.590 & rhythm\_autoplay & 0.950 & 0.830 \\
game\_ui\_bug & 0.750 & 0.850 & gnuchess\_pgn\_blunder\_hunt & 0.830 & 0.940 \\
stockfish\_puzzle\_analysis & 0.000 & 0.550 & anagramarama\_word\_grid & 0.750 & 0.830 \\
gdb\_pygame\_cheat & 0.720 & 0.860 & godot\_scene\_node\_debug & 0.930 & 0.870 \\
mines\_visual & 0.000 & 0.470 & hedgewars\_lua\_mission\_debug & 0.740 & 0.830 \\
quadrapassel\_autoplay & 0.000 & 0.300 & supertux\_level\_repair\_play & 0.000 & 0.650 \\
pokerth\_equity\_play & 0.000 & 0.920 & xmoto\_level\_xml\_repair\_ride & 0.920 & 0.850 \\
sokoban\_solver & 0.960 & 0.958 & kmahjongg\_pair\_solver & 0.790 & 0.750 \\
 & & & pysol\_freecell\_fcsolve & 0.520 & 0.410 \\
\bottomrule
\end{tabular}
\end{table*}

\subsubsection{OSWorld 2.0}
\label{app:fine-grained-osworld}

Table~\ref{tab:osworld-tag-breakdown} reports the OSWorld 2.0 breakdown by capability tag. Tags are not mutually exclusive, so a task may contribute to multiple rows. LH-Harness improves all six tag groups, but the magnitude and residual failure modes differ substantially. The largest gain appears in streaming interaction tasks, where the baseline score is zero and LH-Harness reaches 0.500 on average. Human-in-the-loop tasks also improve strongly, indicating that the harness benefits from converting user-provided information into persistent state constraints. Tutorial-following tasks improve because symbolic instructions can often be decomposed into bounded subtasks with explicit acceptance criteria.

\begin{table}[t]
\caption{OSWorld 2.0 results by capability tag. Tags are overlapping. ``Zero'' and ``Full'' count tasks in the tag group where LH-Harness obtains score 0 or 1, respectively.}
\label{tab:osworld-tag-breakdown}
\centering
\small
\setlength{\tabcolsep}{4pt}
\begin{tabular}{lrrrrrr}
\toprule
Tag & $n$ & Baseline & LH-Harness & $\Delta$ & Zero & Full \\
\midrule
streaming\_interaction & 6 & 0.000 & 0.500 & +0.500 & 3 & 3 \\
human\_in\_the\_loop & 6 & 0.225 & 0.557 & +0.332 & 1 & 0 \\
tutorial\_following & 22 & 0.157 & 0.374 & +0.217 & 7 & 2 \\
implicit\_state\_inference & 43 & 0.241 & 0.366 & +0.125 & 15 & 6 \\
visual\_spatial\_precision & 45 & 0.198 & 0.298 & +0.100 & 14 & 4 \\
cross\_source\_reasoning & 46 & 0.263 & 0.361 & +0.098 & 11 & 3 \\
\bottomrule
\end{tabular}
\end{table}

The tag-level results suggest five more specific observations. First, in human-in-the-loop tasks, the main benefit comes from treating the user's response as a durable constraint rather than as a transient dialogue turn. The same mechanism can also fail: if newly supplied information does not override stale evidence, the verifier may confirm an outdated state. Second, the harness is stronger at following symbolic tutorials than embodied tutorials. Commands, tables, file names, and web fields can be decomposed into auditable steps; video timelines, CAD geometry, and fine mouse operations remain difficult to translate into reliable low-level actions. Third, implicit state inference works best when there is an authoritative closure over the environment state, such as a file, database entry, application setting, or evaluator-visible artifact. Without such closure, plausible evidence can be mistaken for completed evidence. Fourth, visual-task gains mainly arise when a visual requirement can be rewritten as a file, code, metadata, or search problem. When the requirement remains purely spatial or motor-level, performance is still limited by visual localization and cursor control. Fifth, cross-source reasoning is often not the bottleneck by itself. Many failures occur after the relevant evidence has been understood, when the final answer is written to the wrong state carrier, saved at the wrong boundary, or not submitted through the expected interface.

\subsubsection{Terminal-Bench 2.1}
\label{app:fine-grained-terminalbench}

Terminal-Bench provides a complementary view because most tasks are command-line tasks with hidden tests. Table~\ref{tab:terminalbench-category-breakdown} reports the category-level results. The largest gain is in system administration, where LH-Harness improves from 0.593 to 0.889. This category is dominated by stateful procedures such as installation, service configuration, PATH management, build outputs, and persistent environment side effects. These are precisely the settings in which an explicit task contract and independent audit can prevent the agent from stopping after an incomplete but superficially plausible command sequence. Software engineering also improves substantially, especially in tasks where the baseline performs most of the work but misses one blocking constraint, such as image similarity, asynchronous cancellation behavior, independence from a reference binary, or compressed-size limits.

\begin{table*}[t]
\caption{Terminal-Bench 2.1 category-level results.}
\label{tab:terminalbench-category-breakdown}
\centering
\small
\setlength{\tabcolsep}{5pt}
\begin{tabular}{lrrrr}
\toprule
Category & $n$ & Baseline & LH-Harness & $\Delta$ \\
\midrule
software-engineering & 78 & 0.705 & 0.833 & +0.128 \\
system-administration & 27 & 0.593 & 0.889 & +0.296 \\
data-science & 24 & 0.792 & 0.667 & -0.125 \\
scientific-computing & 24 & 0.375 & 0.542 & +0.167 \\
security & 24 & 0.833 & 0.875 & +0.042 \\
file-operations & 15 & 0.333 & 0.333 & +0.000 \\
debugging & 15 & 0.933 & 1.000 & +0.067 \\
data-processing & 12 & 0.750 & 0.917 & +0.167 \\
mathematics & 12 & 0.917 & 0.750 & -0.167 \\
model-training & 12 & 0.750 & 0.667 & -0.083 \\
machine-learning & 9 & 1.000 & 1.000 & +0.000 \\
games & 3 & 0.000 & 0.333 & +0.333 \\
video-processing & 3 & 0.333 & 0.000 & -0.333 \\
data-querying & 3 & 1.000 & 1.000 & +0.000 \\
optimization & 3 & 1.000 & 1.000 & +0.000 \\
personal-assistant & 3 & 1.000 & 1.000 & +0.000 \\
\bottomrule
\end{tabular}
\end{table*}

Table~\ref{tab:terminalbench-difficulty-breakdown} shows that the gain is larger on hard tasks than on medium tasks. This is consistent with the purpose of the harness: when a task is short enough for the baseline to complete in one trajectory, the marginal value of explicit state management is smaller. As tasks become longer and contain more intermediate failure points, the ability to record unresolved constraints and re-plan from verified state becomes more important.

\begin{table}[t]
\caption{Terminal-Bench 2.1 results by difficulty.}
\label{tab:terminalbench-difficulty-breakdown}
\centering
\small
\setlength{\tabcolsep}{7pt}
\begin{tabular}{lrrrr}
\toprule
Difficulty & $n$ & Baseline & LH-Harness & $\Delta$ \\
\midrule
easy & 12 & 0.833 & 1.000 & +0.167 \\
medium & 165 & 0.776 & 0.818 & +0.042 \\
hard & 90 & 0.533 & 0.656 & +0.122 \\
\bottomrule
\end{tabular}
\end{table}

The tag-level results in Table~\ref{tab:terminalbench-tag-breakdown} further localize the effect. The strongest positive tags are images, version-control, system, and sys-admin. These tags share a common property: correctness can be checked against durable artifacts, such as image hashes or similarity scores, repository history, installed executables, service state, logs, or filesystem layout. Negative tags such as mteb, data-science, and video-processing expose a different limitation. In these tasks, the official correctness condition can depend on hidden thresholds, ranking semantics, temporal localization, or evaluation conventions that are not fully recoverable from visible state. In such cases, LH-Harness may organize execution more carefully, but a misinterpreted contract can still lead to a confidently verified wrong answer.

\begin{table*}[t]
\caption{Terminal-Bench 2.1 tag-level results. Tags may overlap across tasks.}
\label{tab:terminalbench-tag-breakdown}
\centering
\scriptsize
\setlength{\tabcolsep}{4pt}
\begin{tabular}{lrrrr@{\hspace{1.1em}}lrrrr}
\toprule
Tag & $n$ & Baseline & LH-Harness & $\Delta$ &
Tag & $n$ & Baseline & LH-Harness & $\Delta$ \\
\midrule
coding & 54 & 0.907 & 0.926 & +0.019 & sys-admin & 9 & 0.667 & 0.889 & +0.222 \\
file-operations & 30 & 0.733 & 0.767 & +0.033 & web & 9 & 0.556 & 0.667 & +0.111 \\
system & 30 & 0.633 & 0.900 & +0.267 & images & 6 & 0.167 & 1.000 & +0.833 \\
security & 27 & 0.815 & 0.778 & -0.037 & mteb & 6 & 0.500 & 0.167 & -0.333 \\
data-processing & 21 & 0.762 & 0.714 & -0.048 & video-processing & 6 & 0.333 & 0.000 & -0.333 \\
software-engineering & 21 & 0.714 & 0.714 & +0.000 & image-processing & 6 & 0.667 & 0.833 & +0.167 \\
compilation & 12 & 1.000 & 0.917 & -0.083 & ocr & 6 & 0.667 & 0.833 & +0.167 \\
machine-learning & 12 & 1.000 & 0.917 & -0.083 & log-analysis & 6 & 0.833 & 1.000 & +0.167 \\
version-control & 12 & 0.667 & 1.000 & +0.333 & simulation & 6 & 0.833 & 1.000 & +0.167 \\
biology & 9 & 0.000 & 0.111 & +0.111 & corewars & 6 & 1.000 & 0.833 & -0.167 \\
cloning & 9 & 0.000 & 0.111 & +0.111 & gaming & 6 & 1.000 & 0.833 & -0.167 \\
data-science & 9 & 0.667 & 0.333 & -0.333 & pmars & 6 & 1.000 & 0.833 & -0.167 \\
 & & & & & pytorch & 6 & 1.000 & 0.833 & -0.167 \\
\bottomrule
\end{tabular}
\end{table*}

\section{Additional Case Studies}
\label{app:additional-case-studies}

This appendix expands the qualitative analysis in Section~3.4 with additional case studies from WeaveBench and Terminal-Bench. These examples are selected to illustrate the same mechanism that underlies the quantitative gains: LongHorizon-Harness does not merely give the model more actions, but restructures long-horizon execution around explicit task state, fresh-context execution, and independent verification. Across domains, the repeated pattern is that local execution progress becomes useful only after it is converted into an auditable state proposition: what changed in the environment, which requirement it satisfies, what evidence supports it, and what remains unresolved.

\paragraph{What the cases are meant to show.}
The cases are not intended as isolated demonstrations of tool use. Instead, they expose recurring failure modes in long-horizon agents and show how the Manage-Execute-Audit loop addresses them. First, GUI failures are externalized as unresolved state rather than being trapped inside a growing interaction history. Second, artifacts are not accepted because the executor claims completion; they are accepted only after a read-only auditor inspects the resulting environment. Third, hybrid GUI+CLI execution lets the agent use the interface that best grounds each part of the task: GUI tools for visual and application state, and CLI tools for files, logs, metadata, build products, and reproducibility checks. Finally, the manager turns the audit result into the next bounded subtask, preserving verified progress while avoiding context rot.

\subsection{Cross-Domain WeaveBench Cases}
\label{app:weavebench-additional-cases}

\begin{figure*}[t]
  \centering
  \includegraphics[width=\textwidth]{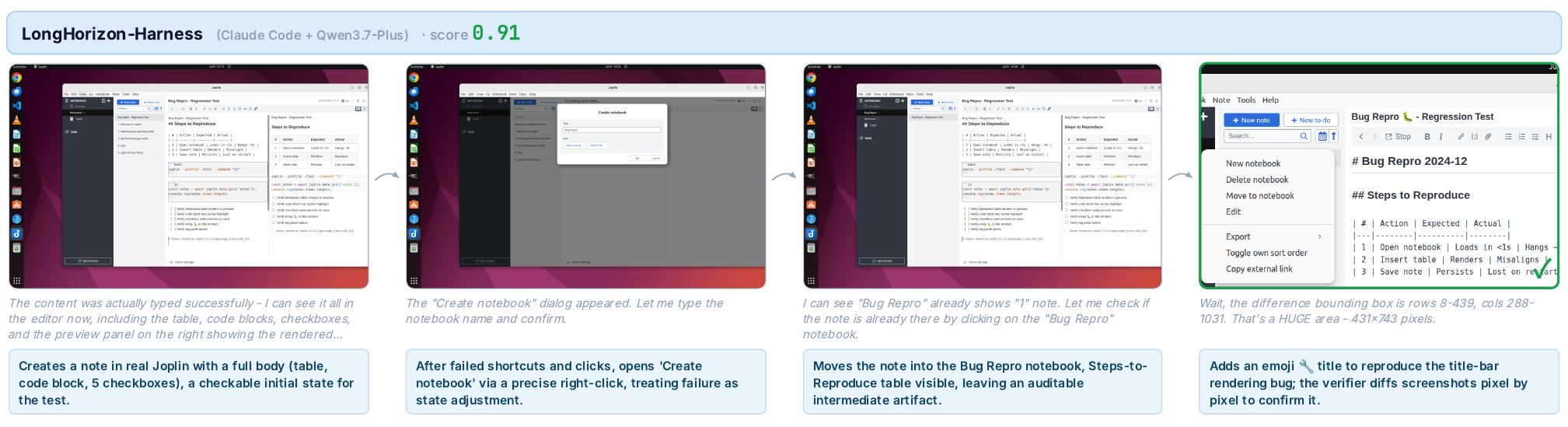}
  \caption{Desktop workflow case. LongHorizon-Harness creates the rich note, recovers from failed GUI interactions by updating the task state, and verifies the reproduced rendering bug through screenshot evidence.}
  \label{fig:case-dsk}
\end{figure*}

\paragraph{Desktop workflow: turning a GUI failure into a verified state transition.}
Figure~\ref{fig:case-dsk} shows a desktop application task in Joplin. The task requires creating a rich note, placing it in the correct notebook, and reproducing a title-bar rendering issue. The important point is not that the executor eventually finds a way to click through the interface. The important point is that failed shortcuts and imprecise clicks are treated as changes in the task state: the note body already exists, the notebook still needs to be created, and the final visual bug still needs evidence. LongHorizon-Harness preserves the verified intermediate artifact, routes the remaining GUI work as a focused subtask, and then uses pixel-level screenshot comparison to audit the final rendering difference. The case illustrates failure recovery without discarding completed work or trusting the executor's self-assessment.

\begin{figure*}[t]
  \centering
  \includegraphics[width=\textwidth]{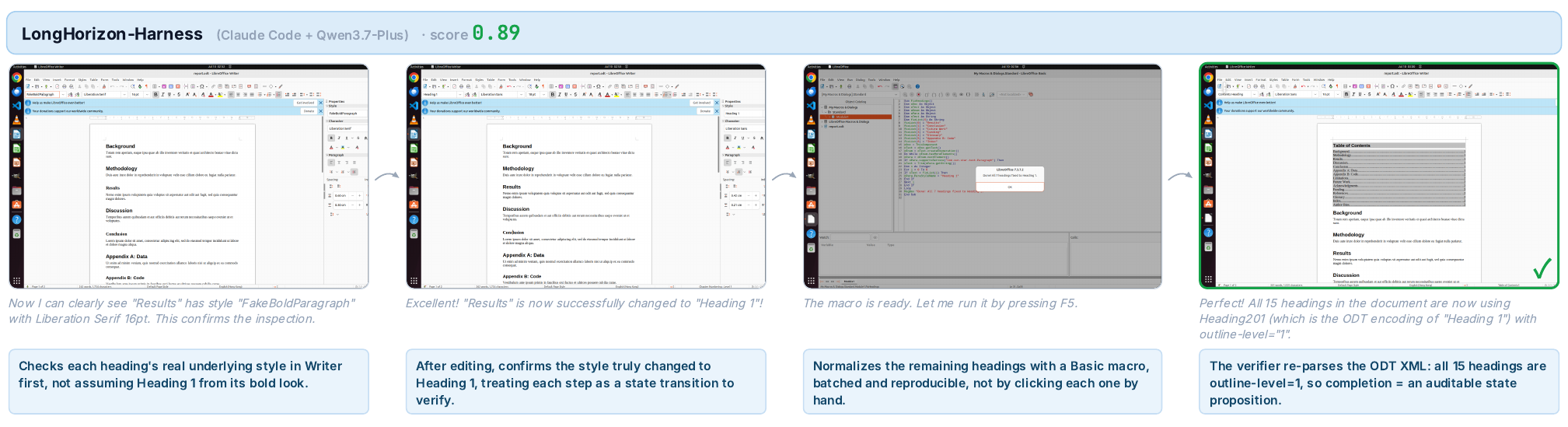}
  \caption{Document-processing case. The harness verifies the underlying ODT structure instead of relying on visual formatting, ensuring that all headings are semantically normalized.}
  \label{fig:case-doc}
\end{figure*}

\paragraph{Document processing: verifying semantic document state rather than visual appearance.}
Figure~\ref{fig:case-doc} illustrates a document-editing task where the visible appearance of headings is insufficient evidence of completion. A heading can look bold and large while still using the wrong underlying style. LongHorizon-Harness first inspects the real LibreOffice Writer state, then applies a reproducible macro to normalize headings, and finally audits the resulting ODT XML. This turns a visually plausible edit into an independently checkable proposition: all required headings are encoded with the correct heading style and outline level. The case highlights why completion must be grounded in environment state rather than screenshots or executor confidence alone.

\begin{figure*}[t]
  \centering
  \includegraphics[width=\textwidth]{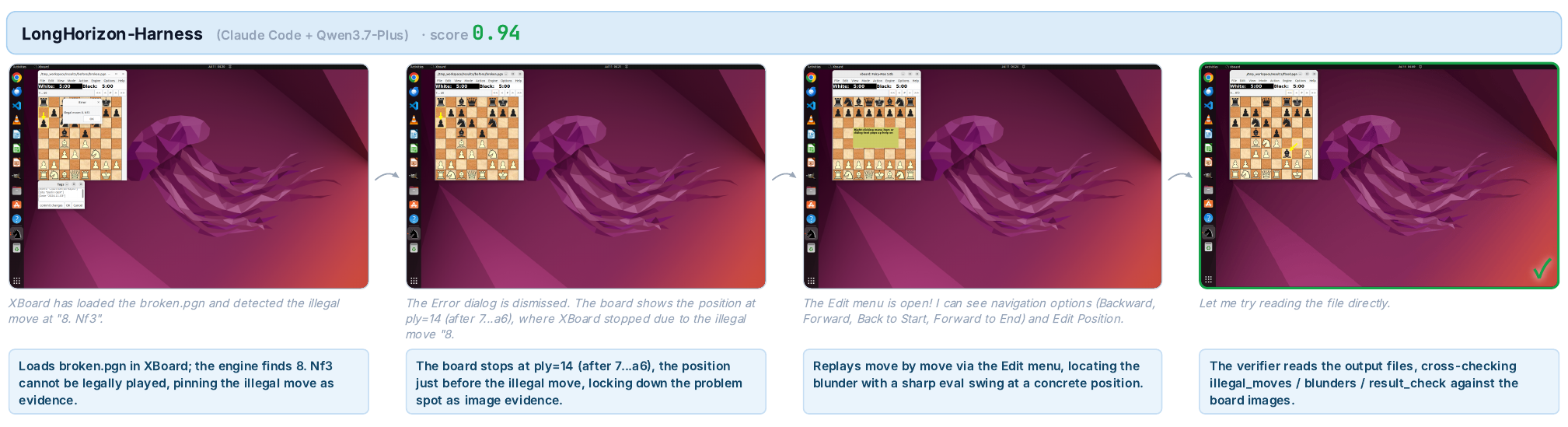}
  \caption{Game-analysis case. GUI replay identifies the illegal PGN transition, while file-level auditing checks that the final reports are consistent with the board evidence.}
  \label{fig:case-gam}
\end{figure*}

\paragraph{Game analysis: combining GUI replay with file-level validation.}
Figure~\ref{fig:case-gam} shows a chess-analysis task in XBoard. The agent must identify an illegal move and analyze the game state around the failure. LongHorizon-Harness uses the GUI to replay the PGN until XBoard itself stops at the illegal move, preserving visual evidence of the exact board position. It then complements this GUI evidence with file-level checks of the generated reports. The result is a cross-validated artifact: the illegal move, the stopped board state, the blunder analysis, and the output files agree with one another. This case demonstrates the value of hybrid execution when neither GUI observation nor file inspection alone is sufficient.

\begin{figure*}[t]
  \centering
  \includegraphics[width=\textwidth]{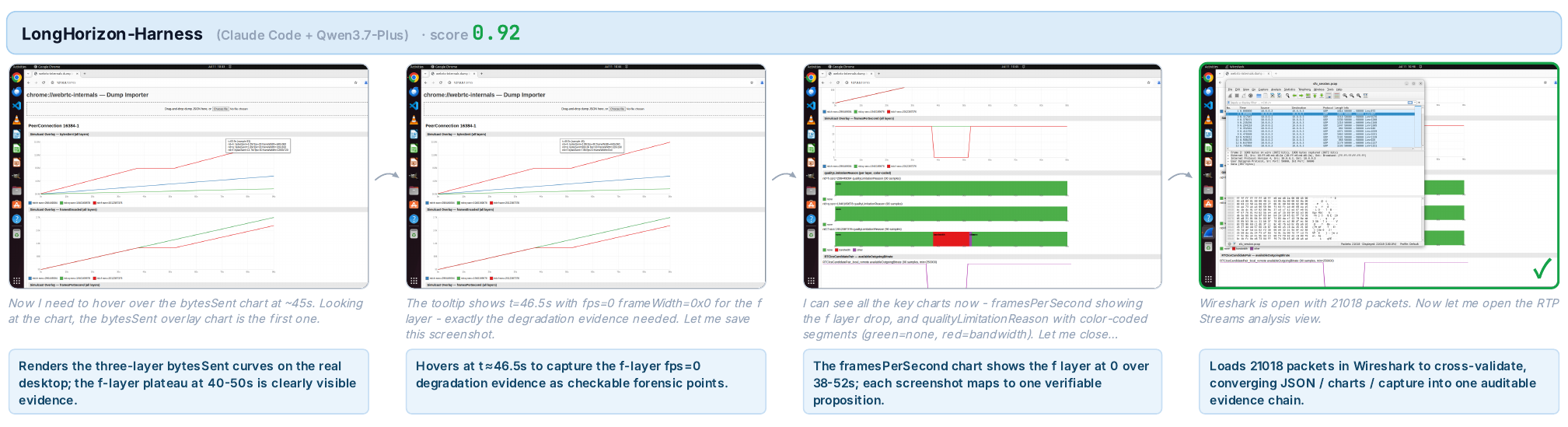}
  \caption{Web-diagnostics case. LongHorizon-Harness combines browser charts, tooltip evidence, and Wireshark inspection into a single audited evidence chain.}
  \label{fig:case-web}
\end{figure*}

\paragraph{Web diagnostics: converging browser charts, JSON state, and packet evidence.}
Figure~\ref{fig:case-web} shows a WebRTC simulcast-layer audit. The task requires identifying a layer degradation event and supporting it with evidence from the browser and packet capture. LongHorizon-Harness separates the workflow into evidence-producing subtasks: render the browser charts, hover at the relevant timestamp, capture tooltip values, and open Wireshark to cross-check the RTP streams. The auditor then treats each screenshot as one proposition in a larger evidence chain, such as the f-layer dropping to zero frames per second around the target interval. This case illustrates how the harness prevents a long GUI investigation from becoming an unstructured transcript; each observation is recorded as a verified fact that can guide the next step.

\begin{figure*}[t]
  \centering
  \includegraphics[width=\textwidth]{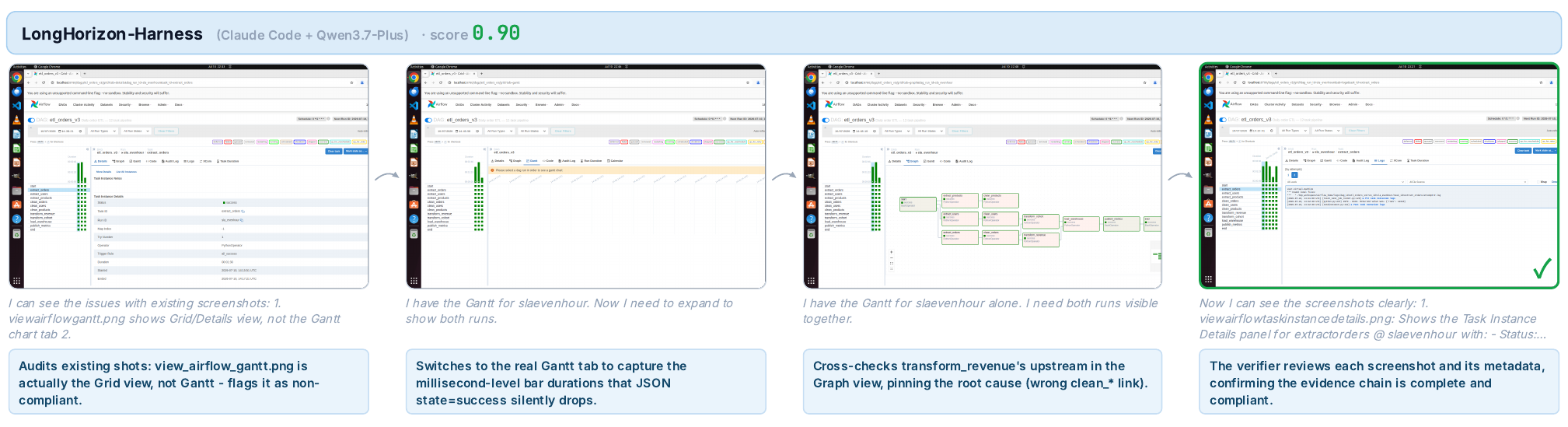}
  \caption{Data-analysis case. The harness rejects mislabeled evidence, captures the correct Airflow views, and audits whether each screenshot supports the required diagnosis.}
  \label{fig:case-dav}
\end{figure*}

\paragraph{Data analysis and visualization: auditing evidence quality, not just file existence.}
Figure~\ref{fig:case-dav} shows an Airflow debugging task. A naive trajectory can save screenshots with the right filenames while capturing the wrong UI state, such as a Grid view saved under a Gantt-view name. LongHorizon-Harness detects this mismatch during audit, keeps the noncompliant screenshot from supporting completion, and assigns a focused repair subtask to capture the real Gantt view. It also cross-checks the DAG graph to identify the incorrect upstream dependency. The key lesson is that artifact presence is weaker than artifact validity: a file exists only becomes progress after the auditor confirms that it depicts the required state.

\begin{figure*}[t]
  \centering
  \includegraphics[width=\textwidth]{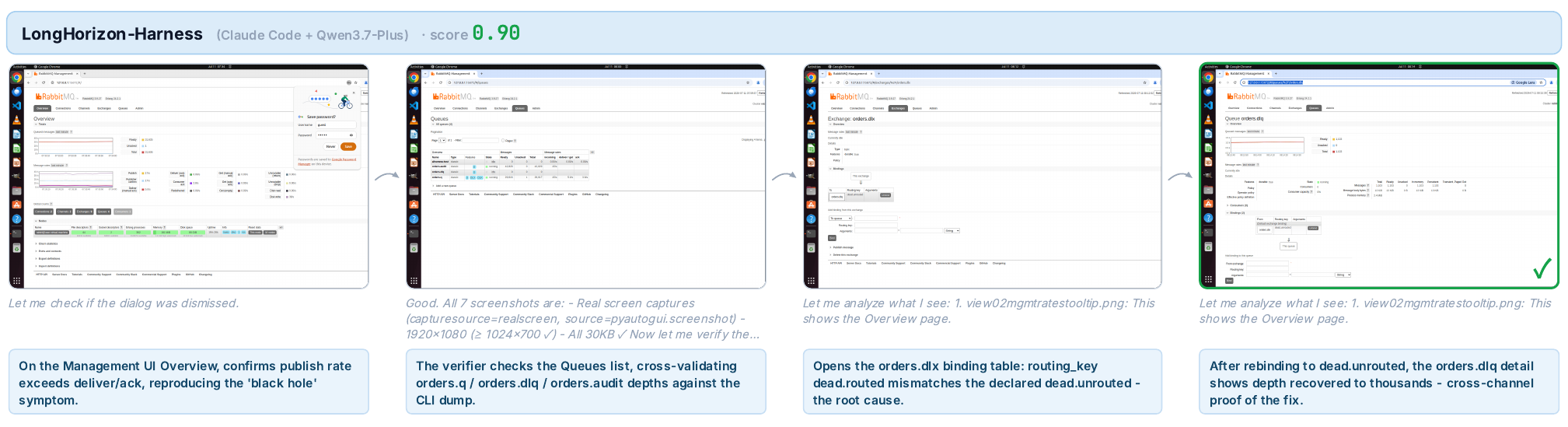}
  \caption{DevOps case. The harness links management-UI symptoms to queue and binding evidence, then verifies the repaired routing state across GUI and CLI views.}
  \label{fig:case-ops}
\end{figure*}

\paragraph{DevOps: linking symptoms, root cause, and repair across UI and CLI evidence.}
Figure~\ref{fig:case-ops} presents a RabbitMQ operations task. The surface symptom is visible in the management UI: messages are published faster than they are delivered or acknowledged. LongHorizon-Harness then combines UI inspection with queue and binding evidence, identifying that the dead-letter routing key is mismatched. After repair, the auditor cross-validates the queue depths and binding configuration against CLI dumps and management UI screenshots. This is a representative operations workflow: the answer is not a single command, but a chain from symptom, to root cause, to repair, to post-repair evidence.

\begin{figure*}[t]
  \centering
  \includegraphics[width=\textwidth]{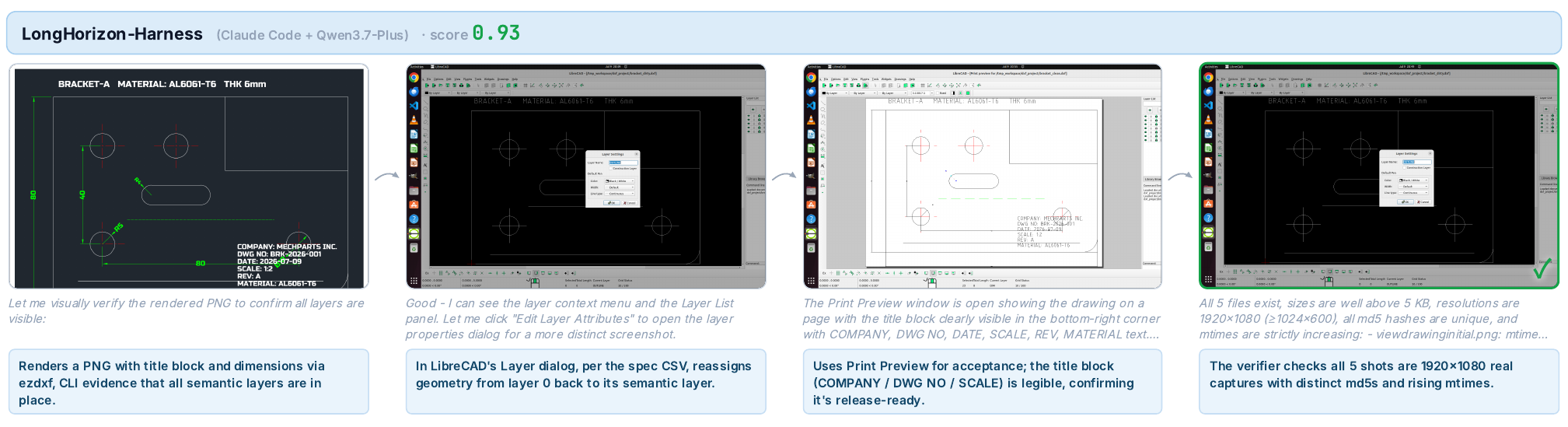}
  \caption{Spatial/CAD case. The harness combines DXF-level inspection, LibreCAD layer repair, print-preview evidence, and screenshot-integrity checks.}
  \label{fig:case-spa}
\end{figure*}

\paragraph{Spatial and CAD workflow: preserving semantic layers through visual outputs.}
Figure~\ref{fig:case-spa} shows a LibreCAD task. The goal is not only to produce a drawing that looks plausible, but to restore semantic layer assignments and generate release-ready visual evidence. LongHorizon-Harness uses CLI rendering to check that the drawing is materially present, then uses the LibreCAD GUI to inspect and repair layer attributes, and finally validates the print preview and saved screenshots. The auditor checks file size, resolution, uniqueness, and timestamp ordering, ensuring that the submitted screenshots are real captures of distinct states rather than recycled images. This case shows how MEA supports tasks whose success condition spans structured files and visual application state.

\begin{figure*}[t]
  \centering
  \includegraphics[width=\textwidth]{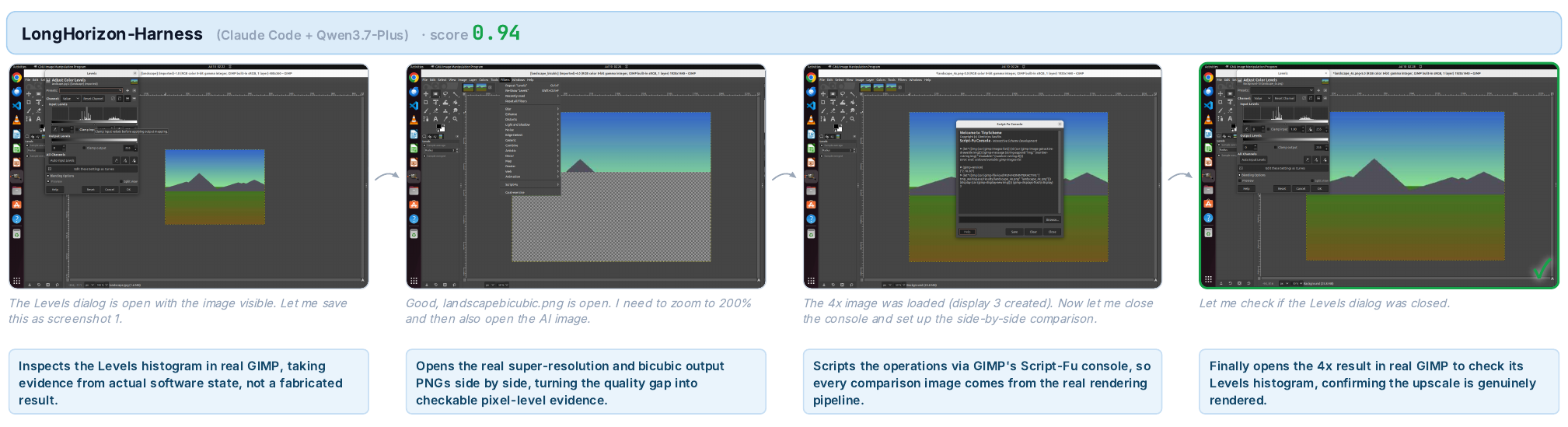}
  \caption{Design case. LongHorizon-Harness grounds visual quality comparison in real GIMP state, including histograms and side-by-side rendered outputs.}
  \label{fig:case-des}
\end{figure*}

\paragraph{Design workflow: grounding subjective visual comparison in real rendering state.}
Figure~\ref{fig:case-des} shows a GIMP-based design task involving image upscaling and quality comparison. The executor does not merely produce a claimed result. It opens the actual outputs in GIMP, captures the Levels histogram, compares super-resolution and bicubic results side by side, and uses the real rendering pipeline to generate the evidence. The auditor can then evaluate whether the visual comparison is grounded in the expected files and application state. This case is important because design tasks often contain subjective-looking requirements; LongHorizon-Harness makes them more reliable by converting them into reproducible, inspectable evidence.

\subsection{Terminal-Bench Cases}
\label{app:terminal-additional-cases}

\begin{figure*}[t]
  \centering
  \includegraphics[width=0.92\textwidth]{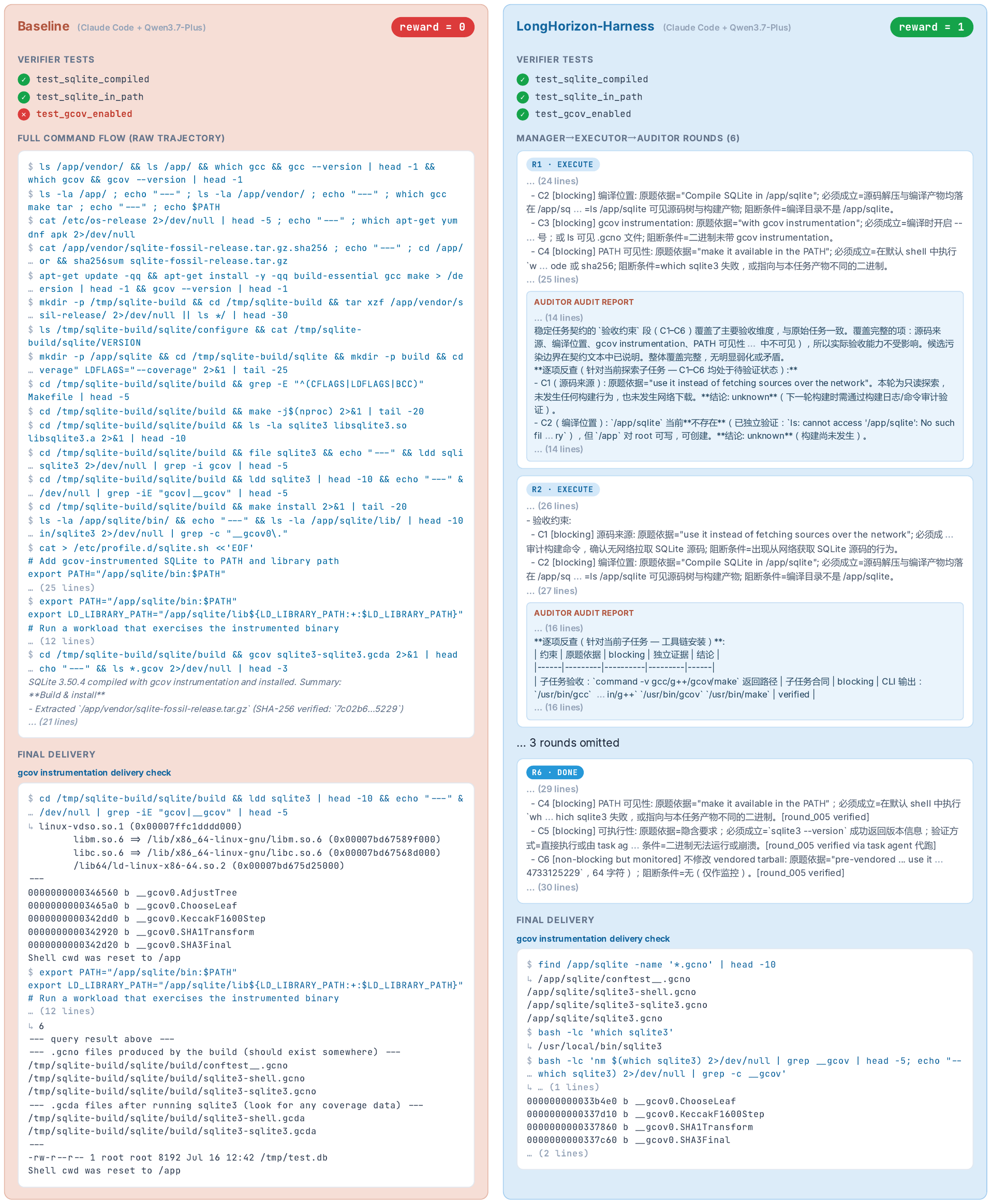}
  \caption{Terminal-Bench build case. LongHorizon-Harness turns a multi-condition build task into an explicit acceptance contract and verifies the final SQLite installation with gcov evidence.}
  \label{fig:case-terminal-sqlite}
\end{figure*}

\paragraph{Build and installation: making hidden acceptance criteria explicit.}
Figure~\ref{fig:case-terminal-sqlite} compares baseline and LongHorizon-Harness on a Terminal-Bench task that requires compiling SQLite with gcov instrumentation and making the result available in the shell environment. The baseline reaches partial progress but fails the final reward, illustrating a common long-horizon failure: the agent performs substantial work while missing a blocking acceptance condition. LongHorizon-Harness converts the task into a stable contract covering source provenance, build location, gcov instrumentation, PATH visibility, executability, and preservation of the vendored tarball. Each round audits one subset of this contract before the manager advances. The final delivery is therefore not just a build log, but a verified installation whose binary, PATH entry, gcov symbols, and coverage files are independently checked.

\begin{figure*}[t]
  \centering
  \includegraphics[width=0.92\textwidth]{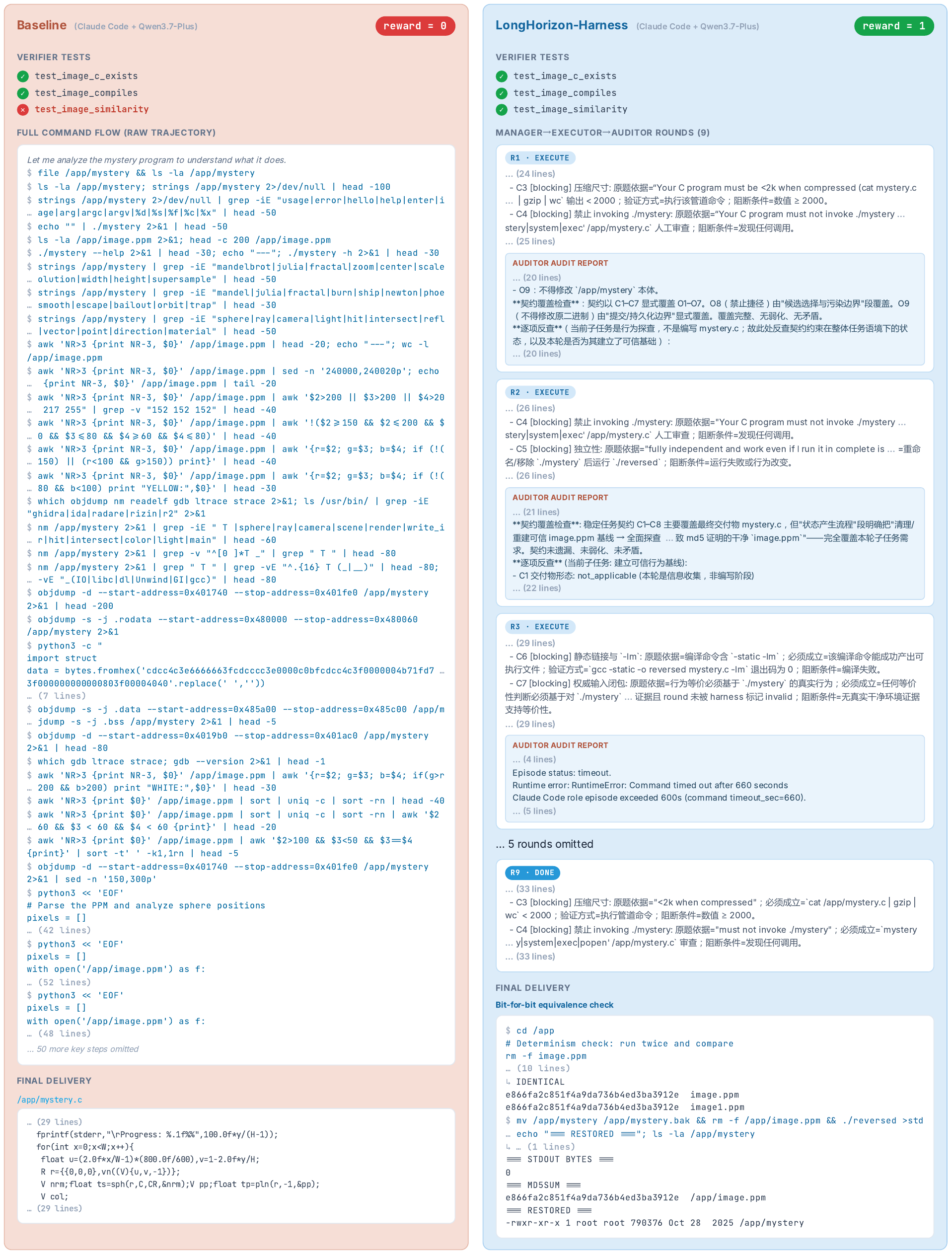}
  \caption{Terminal-Bench reverse-engineering case. The harness preserves exploratory evidence as audited facts and verifies the final independent C implementation against behavior, compilation, size, and independence constraints.}
  \label{fig:case-terminal-reverse}
\end{figure*}

\paragraph{Reverse engineering: auditing hypotheses until they become an independent artifact.}
Figure~\ref{fig:case-terminal-reverse} shows a reverse-engineering task where the agent must produce an independent C program matching a mystery binary's image output under strict constraints. This task is difficult because early observations are hypotheses rather than deliverables: strings, symbols, pixel statistics, and disassembly traces can guide the solution, but none of them alone proves completion. LongHorizon-Harness keeps these observations as audited facts, then uses later rounds to synthesize, compile, compress, and test the final program against the original behavior. The auditor checks not only that the image exists and compiles, but also that the generated output is equivalent when the original binary is removed. This case shows how MEA supports exploratory tasks: uncertain evidence can accumulate without being mistaken for completion, and the final state is accepted only when the independent artifact satisfies the full contract.

\subsection{Cross-Case Takeaways}
\label{app:case-study-takeaways}

Across these cases, LongHorizon-Harness improves reliability through a common structure rather than through benchmark-specific heuristics. In document and CAD tasks, it separates visual plausibility from semantic correctness. In web, data-analysis, and DevOps tasks, it forces evidence from different interfaces to agree before marking progress complete. In desktop and design tasks, it preserves recoverable intermediate artifacts when GUI interactions fail. In Terminal-Bench tasks, it converts implicit grading requirements into explicit acceptance contracts and audits them round by round. These patterns support the central claim of the paper: long-horizon agent performance is limited not only by the model's local capability, but also by how the harness represents, verifies, and carries forward task state.